\documentclass[10pt,journal,compsoc]{IEEEtran}
\newif\ifpeerreview
\peerreviewfalse 
\usepackage[noadjust]{cite}
\usepackage{url}
\usepackage{amsmath,amssymb,graphicx}
\usepackage{lipsum} 
\usepackage[switch]{lineno}
\newcommand{\paperID}{58}
\usepackage{xcolor}
\usepackage{relsize}
\usepackage{placeins}
\usepackage[pagebackref=false,breaklinks=true,colorlinks,bookmarks=false]{hyperref}

\usepackage{pdfpages} 
\usepackage{pgffor} 
\makeatletter
\AtBeginDocument{\let\LS@rot\@undefined}
\makeatother
\def\supplementfilename{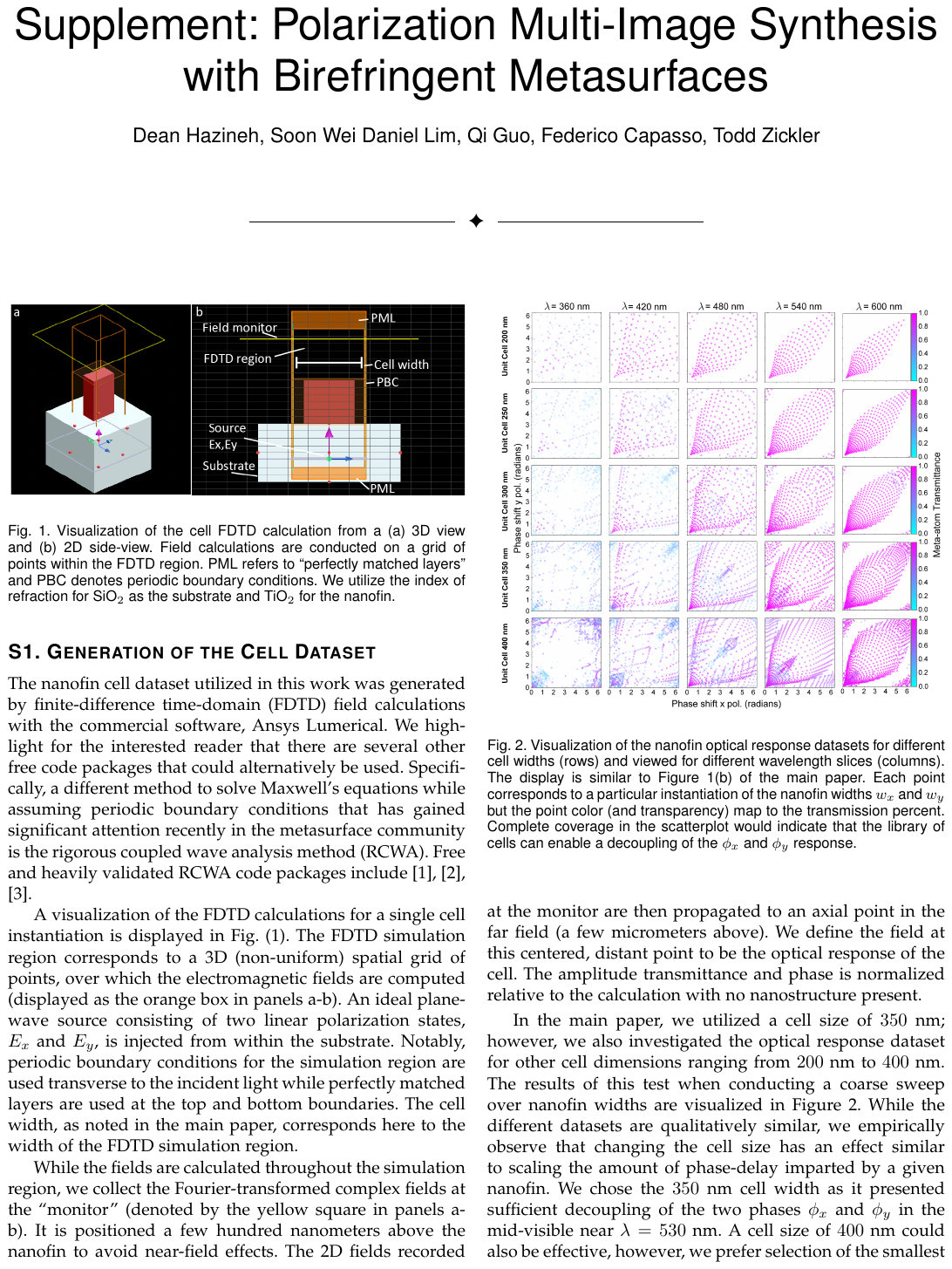}
\pdfximage{\supplementfilename}
\def\numbersupplementpages{\the\pdflastximagepages}

\newif\ifarXiv
\arXivtrue 

\title{Polarization Multi-Image Synthesis with Birefringent Metasurfaces}
\author{Dean Hazineh$^*$, Soon Wei Daniel Lim$^*$, Qi Guo, Federico Capasso, Todd Zickler
\IEEEcompsocitemizethanks{
\IEEEcompsocthanksitem $^*$These authors contributed equally to this work \IEEEcompsocthanksitem D. Hazineh, S.W.D. Lim, F. Capasso are with the Department of Applied Physics, Harvard University, Boston, MA\protect\\
E-mail: dhazineh@g.harvard.edu
\IEEEcompsocthanksitem Q. Guo is with the Elmore Family School of Electrical and Computer Engineering, Purdue University,	West Lafayette, IN
\IEEEcompsocthanksitem T. Zickler is with the Department of Electrical and Computer Engineering, Harvard University, Boston, MA}
}

\usepackage{fancyhdr}
\fancypagestyle{firststyle}
{
   \fancyhf{}
   
   \fancyfoot[C]{\scriptsize This is a preprint of a paper that was presented at the 2023 IEEE International Conference on Computational Photography. The final version is available at IEEE Xplore. © 2023 IEEE. Personal use of this material is permitted. Permission from IEEE must be obtained for all other uses, in any current or future media, including reprinting/republishing this material for advertising or promotional purposes, creating new collective works, for resale or redistribution to servers or lists, or reuse of any copyrighted component of this work in other works.}
}

\begin{document}
\IEEEtitleabstractindextext{%
\begin{abstract}

Optical metasurfaces composed of precisely engineered nanostructures have gained significant attention for their ability to manipulate light and implement distinct functionalities based on the properties of the incident field. Computational imaging systems have started harnessing this capability to produce sets of coded measurements that benefit certain tasks when paired with digital post-processing. Inspired by these works, we introduce a new system that uses a birefringent metasurface with a polarizer-mosaicked photosensor to capture four optically-coded measurements in a single exposure. We apply this system to the task of incoherent opto-electronic filtering, where digital spatial-filtering operations are replaced by simpler, per-pixel sums across the four polarization channels, independent of the spatial filter size. In contrast to previous work on incoherent opto-electronic filtering that can realize only one spatial filter, our approach can realize a continuous family of filters from a single capture, with filters being selected from the family by adjusting the post-capture digital summation weights. To find a metasurface that can realize a set of user-specified spatial filters, we introduce a form of gradient descent with a novel regularizer that encourages light efficiency and a high signal-to-noise ratio. We demonstrate several examples in simulation and with fabricated prototypes, including some with spatial filters that have prescribed variations with respect to depth and wavelength.\\
\newline Visit the Project Page: \url{https://deanhazineh.github.io/publications/Multi_Image_Synthesis/MIS_Home.html}
\end{abstract}
 \begin{IEEEkeywords} 
    Metasurface, Image processing, Polarization-Encoded Point-Spread Functions, Optical Filtering
\end{IEEEkeywords}

}

\maketitle
\thispagestyle{empty} 
\newcommand{\psfA}{\ensuremath{ h_{0^\circ} }}
\newcommand{\psfB}{\ensuremath{h_{45^\circ}}}
\newcommand{\psfC}{\ensuremath{h_{90^\circ}}}
\newcommand{\psfD}{\ensuremath{h_{135^\circ}}}
\newcommand{\psiA}{\ensuremath{ \psi_{0^\circ} }}
\newcommand{\psiC}{\ensuremath{ \psi_{90^\circ} }}

\newcommand{\lnormOne}[1]{\ensuremath{ \left\| {#1} \right\| }}
\newcommand{\lnormTwo}[1]{\ensuremath{ \left\| {#1} \right\|_2 }}
\newcommand\todo[1]{\textcolor{red}{TODO: #1}}
\thispagestyle{firststyle} 

\IEEEraisesectionheading{
\section{Introduction}
}\label{sec:introduction}

\IEEEPARstart{T}here is a rich history in computational imaging of using measurements that are ``coded'', meaning they are recorded by photosensor arrays that are coupled with task-specific, spatially-modulating optics. Multi-shot systems record two or more of these coded measurements sequentially over time, often through dynamic aperture patterns that are implemented by mechanized optics or controllable spatial light modulators. By combining the coded measurements with suitable digital processing, multi-shot systems have played an important role in depth sensing~\cite{pentland1987new, farid1998range, zhou2011coded, Levin2010AnalyzingDF}, wavefront sensing~\cite{Teague1983, Streibl1984PhaseIB, Ichikawa1988}, light field imaging~\cite{liang2008programmable} and hyperspectral imaging~\cite{Kittle2010, August2013, saragadam2022programmable, Salesin2022}. 

Motivated by a desire for improved temporal resolution, there is also work on systems that capture multiple coded measurements in a single exposure. Most of these use a Bayer-like photosensor, which has a pixel-aligned mosaic of three spectral filters, in conjunction with a wavelength-dependent spatial modulator that induces distinct codes on the three channels. Early examples use this approach to acquire depth maps, all-in-focus images, or hyperspectral images~\cite{amari1992single, Bando2008, chakrabarti2012depth,rueda2015multi}. Improvements to functionality and performance have continued, using the conventional three spectral channels (e.g.~\cite{Ikoma2021, baek2021single}) or more spectral channels~\cite{Monakhova20}. 

Analogous to Bayer or spectrally-mosaicked filter arrays, photosensors with interleaved polarization filters are now also quite common~\cite{PolarCam, Pyxis, FLIR}. These measure four linear polarization channels and provide a new avenue for snapshot multi-coded imaging. For example, the recent work of Ghanekar et al.~\cite{Bhargav2022} uses two of the four polarization channels with a task-specific, polarization-dependent spatial modulator for snapshot depth imaging. In our work, we aim to expand the capabilities and potential of multi-coded imaging with polarization.
\begin{figure}[t!]
    \centering
    \includegraphics[width=0.90\columnwidth]{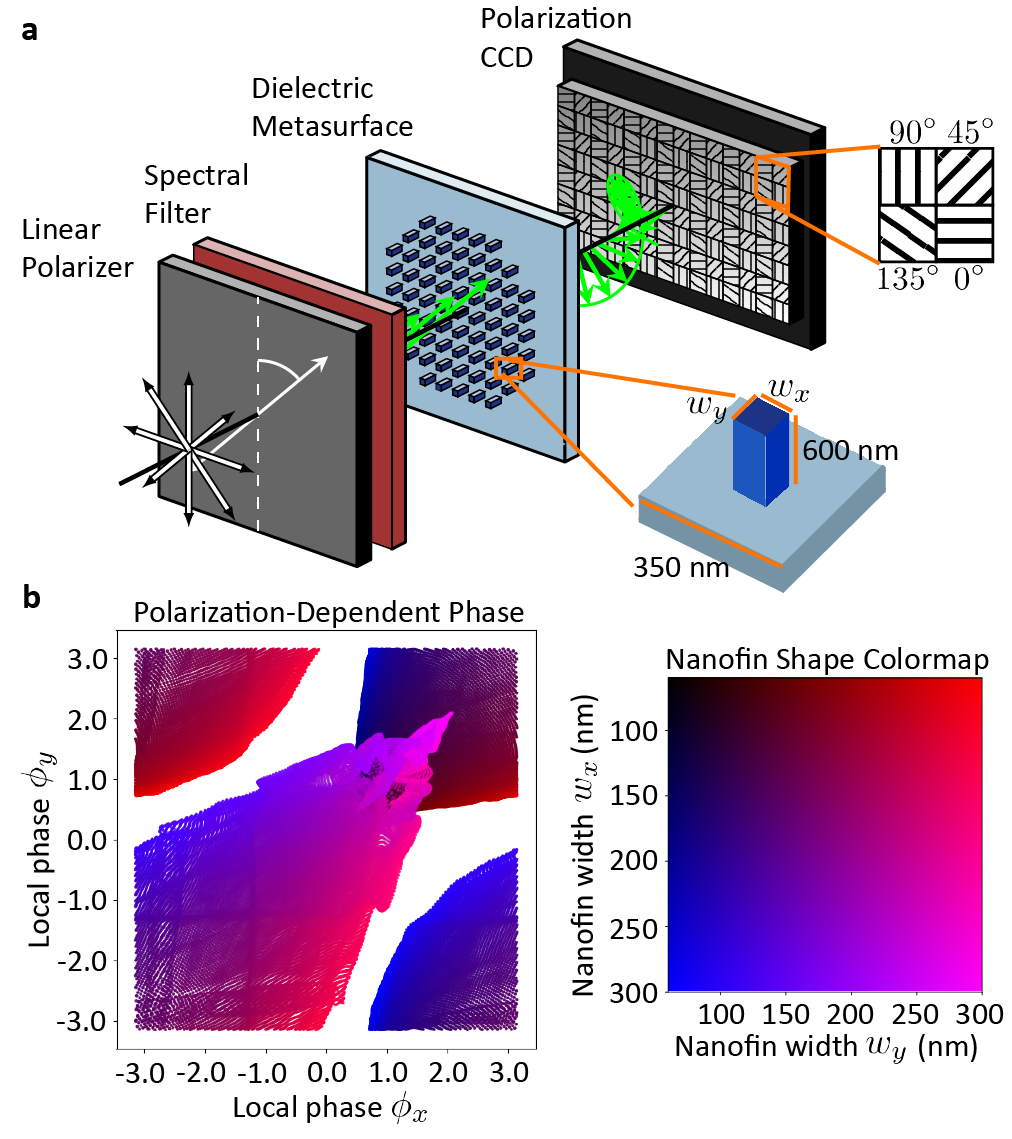}
    \caption{(a) Our system includes a birefringent metasurface and a polarization-mosaicked sensor, optionally preceded (see text) by a standard linear polarizer and narrow-band spectral filter. The metasurface comprises nanofins with varying widths $w_x,w_y$. Each nanofin imparts local phase delays $\phi_x,\phi_y$ (in radians) to two linear polarization states (in addition to amplitude modulations, not shown here). (b) Visualization of the local phase delays imparted by a single nanofin as a function of its widths, as computed by a field solver for incident light of wavelength $532$nm. White areas cannot be imparted by any pair of widths in this range.}
    \label{fig:metalens_schematic}
\end{figure}

Specifically, we explore the design and functionality of a new snapshot system that uses a birefringent metasurface and a polarizer-mosaicked photosensor, as depicted in Figure~\ref{fig:metalens_schematic}a. While there are other polarization-dependent optical components that may be used for spatial modulation at the aperture plane, metasurfaces stand out for their ability to produce distinct, spatially-varying transformations of an incident field for different polarization states~\cite{Rubin21}. We apply our system to the task of opto-electronic filtering, where the digital spatial filtering operation on an image is replaced by the weighted, pixel-wise summation of the four optically-encoded measurements captured on the sensor's four polarization channels. This task is inspired by classical work on optical image processing~\cite{Chavel1976, Lohmann1978}, where a filtered image of a scene is synthesized by the pixel-wise subtraction of two (unpolarized) coded measurements, captured simultaneously using a beamsplitter and distinct modulators placed in parallel optical paths. 

The technical heart of our paper is an approach to solve a related class of computational design problems which we call $\textit{multi-image synthesis}$ problems. In the simplest case, we are given the specification of two real-valued spatial filtering kernels $f^{(1)}(u,v), f^{(2)}(u,v)$, along with the depth $z$ and wavelength $\lambda$ of an ideal axial point source. For these, we aim to design the arrangement of nanostructures on the metasurface such that the spatial-polarimetric interference pattern they induce on the sensor yields four, non-negative per-channel point spread functions (PSFs) $h_c(u,v)$ that can synthesize the specified filters via pixel-wise linear combinations:
\begin{linenomath}
\begin{equation*}
f^{(i)}(u,v; z, \lambda) \approx  \sum_c\! \alpha^{(i)}_c h_c(u,v; z, \lambda),\ c\in \{0^\circ\!,45^\circ\!,90^\circ\!,135^\circ\!\}
\end{equation*}
\end{linenomath}
for some set of digital weights $\alpha^{(1)}_c,\alpha^{(2)}_c\in\mathbb{R}$. 

We solve these problems by using a pre-trained multi-layer perceptron (MLP) to differentiably map the collection of nanostructure shapes, parameterized by roughly $10^7$ total parameters, to their optical responses. We then use gradient descent through a differentiable field propagator to find the set of nanostructures and digital weights that locally minimize the approximation error. In doing so, we find it necessary to introduce a new regularizer that constrains the solution space and encourages the per-channel point-spread functions to be light-efficient, spatially compact, and mutually orthogonal. 

We highlight that, in theory, the four coded measurements captured by the sensor's four linear polarization channels cannot be independently designed, because the specification of two PSFs $\psfA, \psfC$ uniquely determines the others. However, we show experimentally that relaxing the design specification to allow $\psfA, \psfC$ to be merely \emph{close} to their target PSFs over a finite domain provides enough flexibility for $\psfB, \psfD$ to be separately and usefully designed. This observation can be exploited not just for our spatial filtering objective, but for any task that uses linear polarization sensors for snapshot coded imaging. 

Like previous approaches to opto-electronic filtering, our system uses optics to reduce the computational complexity of spatial filtering operations to a pixel-wise summation that is independent of filter size. However, compared to previous approaches it offers several advantages. First, it is compact because it avoids beamsplitters and other bulky refractive elements. Second, by increasing the number of coded measurements from two to four, it can synthesize spatial filtering operations corresponding to any linear combination of two target filters (and thus an infinite set of spatial filtering kernels) by changing only the digital summation weights. Third, the spatial filtering kernels can be designed to match a prescribed depth or wavelength dependence, thereby producing synthesized images that have no equivalent post-capture, digital counterparts. Fourth and finally, by capturing multiple images on distinct polarization channels instead of spectral channels, we can enforce the functionality of the system without introducing assumptions about the scene's material properties. As a result, this is the first compact (single-optic) demonstration of snapshot incoherent image processing suitable for real-world scenes. 

We apply our system to various optical image processing tasks and perform evaluations in simulation and with a prototype camera. Visit the project page\footnote{\url{https://deanhazineh.github.io/publications/Multi_Image_Synthesis/MIS_Home.html}} for more discussion. In addition to providing the code and data for the specific results in this paper, we also create and release a much larger open-source package, called D-Flat, for comprehensive end-to-end metasurface design\footnote{\url{https://github.com/DeanHazineh/DFlat}}. 

We summarize the contributions of this paper as follows:
\begin{itemize}
    \item We propose a metasurface-based architecture to capture four images simultaneously on different polarization channels. Although the measurements are theoretically not independent, we demonstrate that in practice they can all be engineered and utilized.
    \item We introduce a generalization of two-channel opto-electronic filtering to multiple channels and demonstrate that gradient descent with a suitable regularizer can find solutions that operate well under standard imaging conditions.
    \item We design several image-synthesis systems that display new functionalities relative to previous work by virtue of metasurface co-optimization. We present validation for the design theory by comparison to numerical field solvers and experiment.
\end{itemize}

\section{Related Work}
\subsection{Metasurface Optics}
Metasurfaces are a class of recently matured optical devices that consist of sub-wavelength scale structures patterned on a planar, transparent substrate. By judiciously selecting the shape of each nanostructure, the local polarization- and wavelength-dependent optical response can be customized. Moreover, by tailoring the arrangement of nanostructures across the surface, metasurfaces can focus light with high efficiency and can produce structured PSFs that complement downstream computational tasks \cite{Yu2011, Khorasaninejad2016}. Detailed reviews outlining the development and theory of optical metasurfaces can be found in \cite{Genevet17, Kamali2018, Rubin21}. Previous metasurface-based systems for snapshot coded imaging have used panchromatic sensors and have captured their coded measurements by designing the optic to induce their (two or four) distinct measurements at spatially-offset locations on the sensor~\cite{Guo2019, Rubin2019, Colburn2020, Shen2023}. In contrast, our system superimposes its coded measurements at the same spatial location on a sensor, and it uses the sensor's polarization mosaic to separately sample them.

\subsection{Neural Representations}
In Section~\ref{sec:neural_model}, we introduce an MLP to efficiently model the mapping from a nanostructure's shape to the optical modulation it imparts on an incident field. This builds on a history of applying deep learning to tasks in nano-photonics, as reviewed in~\cite{Wiecha2021, Jiang2021}. Most similar to us are uses of fully connected neural networks for mapping shape to broadband phase~\cite{Jiang21, Sensong2019, Peurifoy2018, Liu2018, Nadell19}. However, our work differs by using neural models in an end-to-end optimization framework, which is reflected in differences in our architecture. Besides predicting the phase and transmittance for two polarization states, we include wavelength as an input to our MLP which provides a low-dimensional input/output mapping that is similar to coordinate-MLPs~\cite{mildenhall2020}.

\subsection{Incoherent Spatial Filtering}
\label{ssec:prior_filtering}
Opto-electronic filtering with incoherent light has recently been revisited in~\cite{Wang2020, Zhang2021}, where a photonic crystal slab or a multi-layer film is coupled with a refractive lens. In both cases, the optical responses at two narrow wavelength bands are engineered to create two coded measurements that are captured at the photosensor using an array of spectral filters. These two types of optical modulators work by imparting a transmission that is dependent on the angle of incidence, and because of this, they can only reshape the Gaussian PSF of the refractive lens and cannot produce more general PSFs like we show in this paper. Moreover, these methods require that all objects in the scene emit light of equal intensity at the two designed wavelength bands, which cannot be enforced in practice and limits their utility. 


\subsection{Constrained Matrix Factorization}
The optimization task that we encounter in this paper is loosely related to prior work on finding constrained matrix factorizations. Specifically, an optimization problem that is related to our main objective is
\begin{equation}\label{eq:semi-nmf}
    \operatorname*{argmin}_{H\ge0,A} \left\| F-HA\right\|^2, 
\end{equation}
where the columns of $F\in\mathbb{R}^{N\times 2}$ are a pair of spatially-discretized target filters to be realized by synthesis. The four columns of $H\subset\mathbb{R}_{\geq0}^{N\times 4}$ are the four (non-negative) component PSFs produced by the optical system and captured at the photosensor, and the two columns of $A\in\mathbb{R}^{4\times 2}$ are the sets of digital image weights. 
Objective~(\ref{eq:semi-nmf}) has been called semi-nonnegative matrix factorization or semi-NMF~\cite{ding2008convex}.

In contrast to us, previous work has explored problems of this form for situations where the columns of $F$ outnumber the columns of $H$, and where the recovered $H$ and $A$ provide clustering or dimensionality reduction. In that context, one usually iterates between updates of $H$ and $A$;  see~\cite{wang2012nonnegative} for an early review. In our case, we use gradient descent because it allows for the incorporation of conditions that are specific to our domain, namely that the columns of $H$ are nonlinearly parameterized by the metasurface shapes and outnumber the columns of $F$; and that neither nonnegativity nor orthogonality constraints are applied to weights $A$. 

\section{Proposed Method}
In this section, we present a method to solve the optimization problem described in the introduction. In doing so, we rely on the principle of incoherent image formation based on the point-spread function (PSF). A simple model follows from imagining a scene to be composed of planar, emitting surfaces at various depths, which are each parallel to the image sensor and do not occlude each other within the field of view. For a polarization channel denoted by $c$, the spectrally-integrated intensity distribution in that channel at the photosensor plane $I_c(u, v)$ can be approximated by the spectral sum of 2D spatial convolutions between the depth-dependent and wavelength-dependent PSF $h_c$ and the (magnified) scene radiance $\mathcal{I}_c$ via
\begin{equation}
     I_c(u,v) = \mathlarger{\sum}_{\lambda} \mathcal{I}_c(u,v,z;\lambda) \underset{(u,v)}{*} h_c(u,v,z; \lambda).
     \label{eq:rendering_eq}
\end{equation}
From the linearity of convolution, it is clear that a pixel-wise linear combination of such measurements $\sum_c\alpha_cI_c$ is equivalent to spatially filtering the scene radiance with an effective ``net PSF'' given by $\sum_c\alpha_ch_c$. 
In what follows, we use polarization channels to capture measurements $I_c$, and so we assume that the scene emits light that is unpolarized, meaning $\mathcal{I}_c = \mathcal{I},\ \forall c $. 
In practice, we can ease this assumption by placing a linear polarizer at the entrance of the optical system, as shown in Figure~\ref{fig:metalens_schematic}a. The relative orientation of the polarizer is chosen to project equal intensity on two specific linear polarization states.  

\subsection{Metasurface Point Spread Function}
\begin{figure}[t!]
    \centering
    \includegraphics[width=0.95\columnwidth]{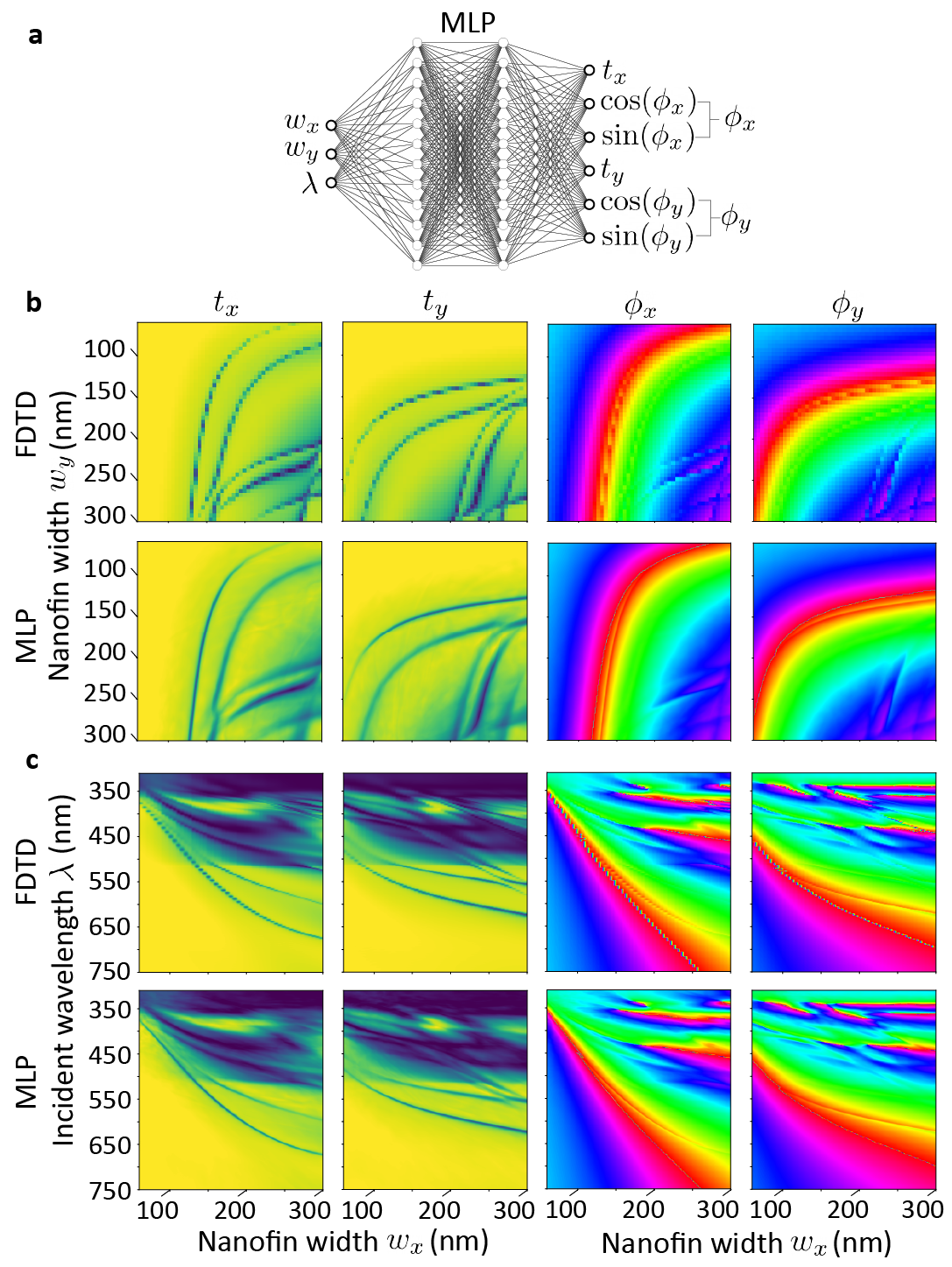}
    \caption{(a) A pre-trained MLP provides an efficient, differentiable proxy for the nanofin field solver (FDTD). It maps shape parameters and incident wavelength to phase and transmittance values for two polarization states. Phase is wrapped to $2\pi$ as drawn. (b,c) Comparisons between FDTD and MLP outputs at 5x the resolution used for pre-training, for (b) fixed wavelength $\lambda=532$ nm and (c) fixed nanofin width $w_y=180$ nm.}
    \label{fig:nanofin_MLP}
\end{figure}

In this work, we define the relationship between the metasurface and the point-spread function $h_c$ by employing a standard cell-based treatment, whereby the metasurface is considered as the composition of smaller building blocks \cite{Yu2011}. While summarized here, a detailed review of the design theory can be found in \cite{Hu2021}. 

We define the metasurface $\Pi$ as a collection of cells on a regular grid of points $\chi$. The nanostructures in each cell may then be specified using a set of shape parameters $\pi$. Here, we consider $350$ nm wide square cells that each contain a single $600$ nm tall nanofin, parameterized by the fin widths $w_x$ and $w_y$, i.e.,
\begin{equation}\label{eq:cellParameter}
    \Pi = \{\pi(x', y') | (x', y') \in \chi \}; \ \pi(x', y') = (w_{x'}, w_{y'}).
\end{equation}
We use an electromagnetic field solver to compute solutions to Maxwell's equations and create samples of the mapping $O$ from the cell to its \textit{local} optical response, given by the wavelength-dependent amplitude transmittance $t_c$ and phase delay $\phi_c$ imparted to an incident wavefront,
\begin{equation}
    O\left(\pi(x', y'), \lambda \right) = t_c(x', y') e^{i\phi_c(x', y')}.
\label{eq:optical_mapping}
\end{equation} 
We then approximate the phase and transmittance profiles of the full metasurface by stitching together the spatial grid of per-cell responses. 

Notably, since the cells are sub-wavelength, its optical response should in fact be dependent on the nanostructures present in neighboring cells. To enable the treatment of a cell as an independent building block, however, a key assumption that is made in the design theory is the application of periodic boundary conditions when solving for the field. By utilizing periodic boundary conditions, we obtain an \textit{approximation} to the true local optical response that is independent to the selection of cells at other locations on the metasurface. In practice, it is observed that this assumption is sufficiently accurate to describe composite, aperiodic devices as long as the spatial gradients $\nabla \pi$ are generally small. In supplement S2, we validate this treatment by designing reduced-size versions of the metasurfaces presented in the results section. We compare the optical response and the PSF obtained when solving for the full field across the metasurface to that obtained when utilizing the cell approach and find close agreement.     

\begin{figure}[t!]
    \centering
    \includegraphics[width=0.9\columnwidth]{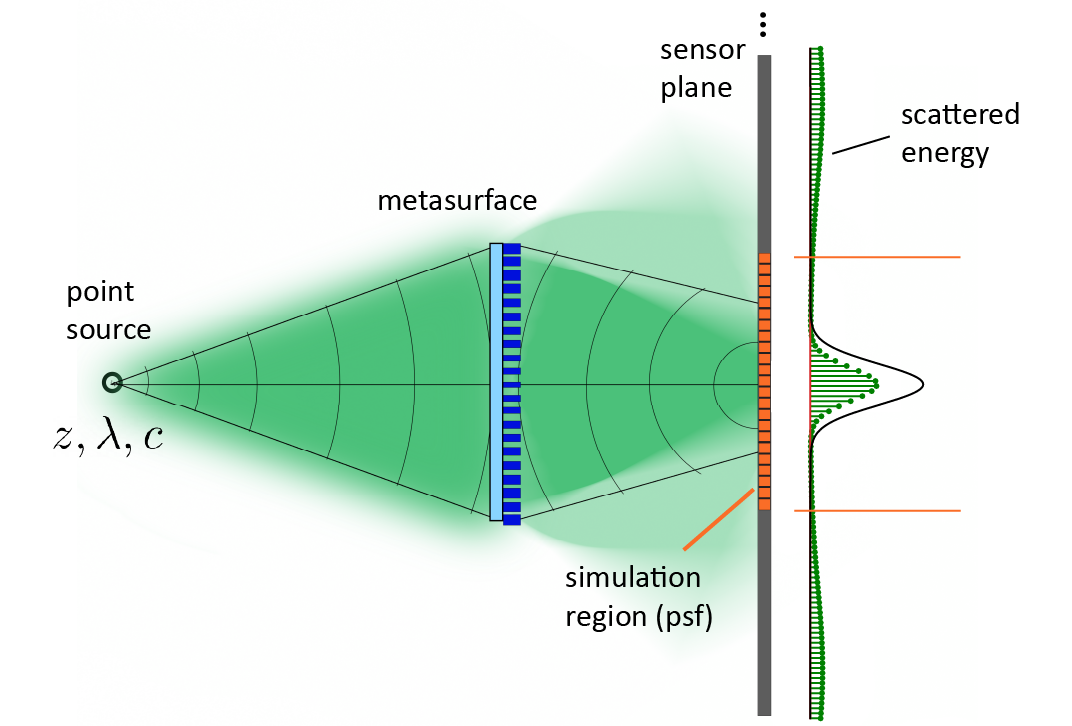}
    \caption{Qualitative depiction of the PSF calculation, shown in 1D. The intensity distribution at the sensor plane (green stems) is computed over a finite simulation region and is optimized to match a target signal (black line) over that domain and only up to scale. Energy that is scattered outside of the calculation region is considered lost and reduces the focusing efficiency of the metasurface.}
    \label{fig:simulation_region}
\end{figure}

We compute the mapping in Equation~(\ref{eq:optical_mapping}) for nanofins made of titanium dioxide ($\text{TiO}_2$) using a commercial finite-difference time-domain (FDTD) solver, assuming normally incident light of two orthogonal polarization states ($0^\circ, 90^\circ$), chosen to be aligned with the $x$ and $y$ axis of $\chi$. The optical response need only be computed for a pair of orthogonal linear polarization states since the response for all other in-plane incident angles may be obtained by a change of basis. More details of the simulation are provided in supplement S1. We sweep nanofin widths between $60$ and $300$ nm, resulting in a dataset of $2304$ cell instantiations, and compute the optical response for wavelengths between $300$ and $750$ nm. Slices from this dataset are displayed in Figure~\ref{fig:nanofin_MLP}b-c.

This set of optical responses constrains the space of possible polarization- and wavelength-dependent PSFs that can be produced by the metasurface. For incident light of a single wavelength $\lambda=532$ nm, we show in Figure~\ref{fig:metalens_schematic}b that the local phase delays, $\phi_x, \phi_y$, imparted to the two polarization states approximately span the full range (wrapped to 2$\pi$) and can be nearly decoupled. We may then consider that a metasurface assembled from a collection of these cells can be used to realize two distinct, spatially varying phase modulations and can produce a pair of PSFs that can be (approximately) independently designed. 

Given the phase and transmittance defined across the metasurface (applied linearly to an incident, spherical wavefront originating from an axial point-source), we obtain the complex PSF at the photosensor a distance $d$ after the optic by per-channel propagation using the Fresnel diffraction equation \cite{Goodman2017}, given in integral form via,
\begin{equation}
    \begin{split}
    \sqrt{h_c(u,v;z)}e^{i\psi_c(u,v;z)} &= \iint T_c(x,y) Q(u,v;x,y) \text{d}x\text{d}y\\
    \end{split}
    \label{eq:propagation}
\end{equation}
where $T_c$ corresponds to the wavefront after the metasurface and $Q$ is the standard Fresnel kernel,
\begin{equation}
    \begin{split}
        T_c(x,y) &=  t_c(x,y)e^{i\phi_c(x,y)} \frac{e^{ikr}}{r} \text{ for } r = \sqrt{x^2 + y^2 + z^2}\\
        Q(u,v;x,y) &= \frac{e^{ikd}}{i\lambda d} \exp\left[\frac{ik}{2d}\left((x-u)^2 + (y-v)^2\right)\right].
    \end{split}
    \label{eq:propagation2}
\end{equation}

In carrying out this calculation, we define a finite \emph{simulation region} $S\subset\mathbb{R}^2$ at the sensor plane, comprised of a uniform grid of points centered around the optical axis. Due to computational constraints, this region covers an area that is smaller than the actual dimensions of the intended photosensor (see Figure~\ref{fig:simulation_region} for a qualitative diagram). Notably, light that is scattered outside of the simulation region is undesirable as it reduces both the contrast and the signal-to-noise ratio of images formed by the system. To quantify the amount of light that is deflected away, we evaluate as a metric the \emph{focusing efficiency}, which is defined as the fraction of incident light on the metasurface that is transmitted and scattered within the simulation region. In the remainder, we use the shorthand $h_c$ to denote the intensity and $\psi_c$ the phase of the field that is induced on the simulation region. 

\subsection{Interference of Birefringent PSFs}
\label{sec:psf_interference}
\begin{figure}[t!]
    \centering
    \includegraphics[width=1.0\columnwidth]{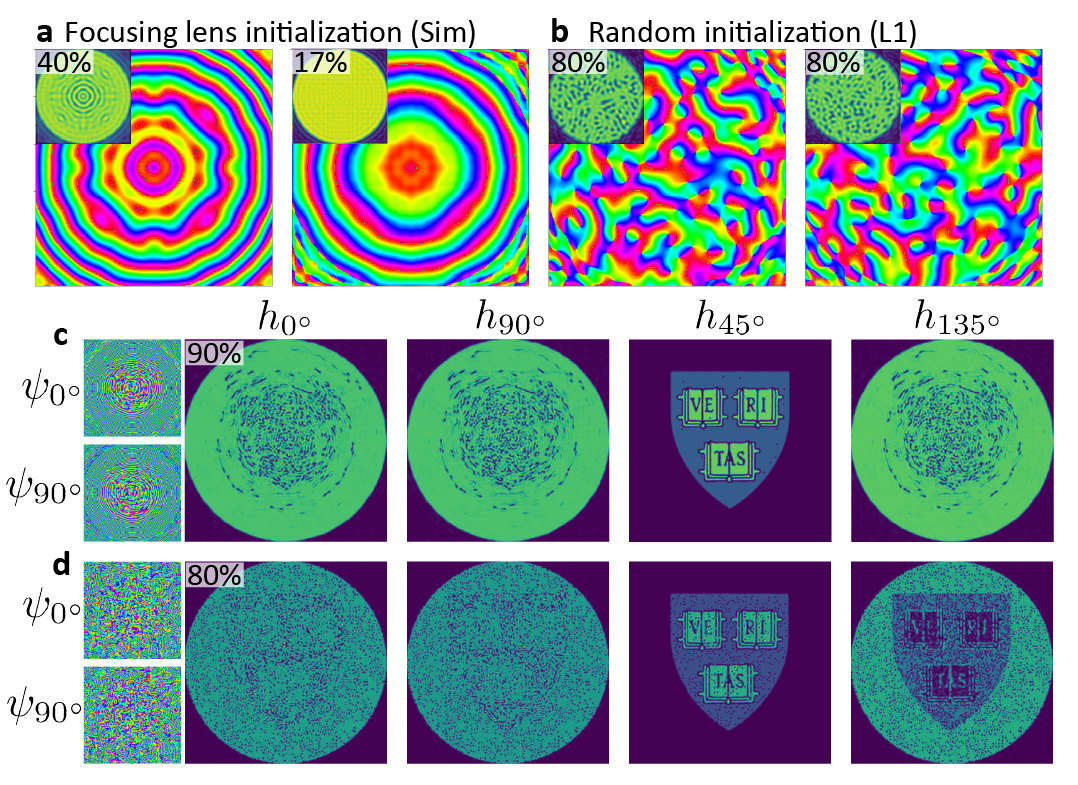}
    \caption{Visualization of the phase $\psiA$ and intensity $\psfA$ (insets) at the photosensor plane for a phase-only optic optimized according to Equation (\ref{eq:phase_inversion}). The target intensity distribution was a uniform disk. A cosine-similarity (sim) loss function was used in (a) while the L1 loss was used in (b). The text percent denotes the focusing efficiency for that particular solution. Different intensity approximations to the target distribution, and thus different output phase distributions, can occur by errors in the intensity within the simulation region or by discarding energy outside of the simulation region. Using a focusing-lens initialization (c) and a random phase initialization (d), a pair of phase profiles for the optic are optimized to approximate specified intensity distributions on $\psfA, \psfC$ (uniform disk) and on $\psfB$, which is formed from the interference.}
    \label{fig:interference_probe}
\end{figure}

While four polarization states are simultaneously sampled by the polarizer-mosaicked photosensor, we note that the set of intensity measurements captured in our system are not fully independent. Specifically, given the intensity and phase of the $0^\circ$ and $90^\circ$ polarized fields that are induced at the sensor plane by the metasurface, the intensity measured on the $45^\circ$ and $135^\circ$ channels may defined in terms of the interference,
\begin{equation}
    h_{45^\circ (135^\circ)} = \frac{h_{0^\circ}}{2}  + \frac{ h_{90^\circ}}{2} \mp \sqrt{h_{0^\circ}h_{90^\circ}}\cos(\psi_{0^\circ} - \psi_{90^\circ}).
    \label{eq:interf_psf}
\end{equation}
Consequently, while the intensity patterns on all four channels are distinct, the design space is constrained and only three measurements are linearly independent. Despite this fact, it is still beneficial to utilize all four measurements as is discussed in Section \ref{sec:MIS_optimization}.

Since our proposed method relies on the ability to engineer the collection of PSFs, we raise the following question: When the intensity distributions $\psfA$ and $\psfC$ are fixed, what is the space of functions that can be realized for $\psfB$ by structuring the phases at the sensor plane, $\psi_{0^\circ}$ and $\psi_{90^\circ}$? For simplicity, let us consider the transmittance of the metasurface to be a uniform disk with a spatial constraint set by the aperture.  In an exact sense, the answer is then disappointing. Transport of intensity~\cite{Chao2020} tells us that specifying the intensity $\psfA$ (or $\psfC$) everywhere on the sensor plane determines the phase $\psiA$ (or $\psiC$), and so the number of possibilities for PSF $\psfB$ is exactly one. 

However, a key concept in this work is that substantially more flexibility emerges in the solution space when we are only interested in (and capable of realizing by gradient descent) intensity distributions at the sensor that $\textit{approximate}$ a target distribution over a finite subset of that plane. Fortunately for us, there are an infinite number of these approximations, and because each corresponds to a different phase distribution, we may control the intensity measured on the interference channels. For our particular task, we also highlight that our approach relies less on having exact intensity distributions for each of the component PSFs and much more on the accuracy of their linear combination. 

To visualize this flexibility, we first borrow inspiration from~\cite{Mahendran2015} and compute different instantiations of the phase at the sensor plane $\psiA$ that emerges when using gradient descent to optimize the intensity $\psfA$ to approximate a target intensity $h'$. We use different initial conditions and terminate descent after a fixed number of steps. Specifically, we solve the following minimization problem to recover the phase modulation on the optic, 
\begin{equation}
    \phi^* = \operatorname*{argmin}_\phi \left[ \mathcal{L}\left( \lvert P(te^{i\phi})\rvert^2 ,h' \right)  \right],
    \label{eq:phase_inversion}
\end{equation}
where P denotes the free-space propagation operator of Equation (\ref{eq:propagation}), transmittance $t$ is set to be unity within an aperture radius, and we consider different loss functions for $\mathcal{L}$ to emphasize qualitatively different intensity solutions. Examples of the optimized sensor plane intensity and phase distributions (produced after propagating the field of $t$ and $\phi^*$) are shown in Figure~\ref{fig:interference_probe}a-b. In panels c-d, we provide a similar visualization demonstrating how these different approximations to intensity enable the ability to uniquely structure the interference. While the pair of intensities $\psfA$ and $\psfC$ are again optimized to approximate the target $h'$, a different user-defined intensity distribution for $\psfB$ can be realized. 

\subsection{Neural Optical Model}
\label{sec:neural_model}
Motivated by the recent success of coordinate-MLPs as neural implicit representations for a suite of tasks \cite{mildenhall2020, tancik2020}, we employ a pre-trained MLP as a differentiable proxy function for the mapping between nanofin cells and their optical response (Equation \ref{eq:optical_mapping}). We consider the network depicted in Figure~\ref{fig:nanofin_MLP}a, consisting of two hidden layers, ReLU activation, and between 256 and 1024 neurons per layer. Min-max normalization is applied to the inputs and phase-wrapping is handled by predicting the geometric projection of the phase (often referred to as cyclical feature encoding). After training on the FDTD data, we find that the model can accurately reproduce the mapping, with a mean absolute error in complex field predictions for a withheld test set as low as 0.019. Qualitatively, we also observe that the model can correctly identify the cells that experience resonances which are characterized by dips in the transmission. The FDTD and MLP outputs are visually compared in Figure \ref{fig:nanofin_MLP}b-c. 

As a benchmark, we compute the number of floating point operations for an equivalent calculation utilizing the auto-differentiable field solver in \cite{colburn2021}. We find that the MLP requires approximately a factor of $10^3$ to $10^4$ fewer floating point operations per evaluation. Additional details are provided in supplement S3. In the supplement, we also compare the usage of an MLP to alternative models including elliptic radial basis function networks and multivariate polynomial functions (as was used for nanocylinder metasurface design in \cite{tseng2021neural}). We find the MLP to be substantially more accurate and to be the only model tested that reproduces the high-spatial frequency features in the data.  

Once trained, the network weights are fixed and the MLP is used for the main optimization tasks in this work. In order to constrain the learned nanofin dimensions $w_x,w_y$ to be within the min-max bounds of the training dataset, we use reparameterization \cite{park1975transformation, Chen2020} and optimize over an unconstrained latent variable that is transformed to the bounded nanofin widths.     

\subsection{Multi-Image Synthesis Optimization}
\label{sec:MIS_optimization}

In this section, we discuss our optimization algorithm to design a metasurface that produces four polarization-encoded measurements for image processing. Since the formation of an incoherent image may be modeled by convolution with the intensity PSF (Equation~\ref{eq:rendering_eq}), spatial frequency filtering objectives may be formulated as the realization of a discretized, target filtering kernel $F \in \mathbb{R}^{B \text{x} N \text{x} 1}$ from the linear combination of non-negative PSFs. Throughout, $B$ denotes a batch dimension corresponding to the channels of wavelength and depth, and $N$ denotes the number of sensor/image pixels used to define the kernel (flattened to 1D).
   
Considering the polarizer-mosaicked photosensor, our optical system is characterized by the collection of four PSFs, defined as $H=[\psfA, \psfB, \psfC, \psfD]$ where $H\subset \mathbb{R}_{\geq0}^{B \text{x} N \text{x} 4}$. The PSFs $\psfA$ and $\psfC$ are computed utilizing Equations (\ref{eq:optical_mapping})-(\ref{eq:propagation2}) for a given metasurface, while $\psfB$ and $\psfD$ are defined according to the interference via Equation (\ref{eq:interf_psf}). A set of digital weights are defined as $\alpha \in \mathbb{R}^{4\text{x}1}$ such that the (noiseless) synthesized net PSF approximating the target filter is given by the tensor product $H\alpha$. Throughout this paper, we use the notation $XY$ to represent batched matrix multiplication between tensors $X$ and $Y$\footnote{More generally, matrix operations applied to a tensor corresponds to the operation on the matrix in the inner-two dimensions. For example, if $H$ has the shape $[B\text{x}N\text{x}4]$, then $H^T$ takes the shape $[B\text{x}4\text{x}N]$.}. The primary task is then to identify suitable decompositions for $\alpha$ and the physics-constrained tensor $H$ (produced by a metasurface $\Pi$) given one or more target filtering kernels.

\begin{figure}[t!]
    \centering
    \includegraphics[width=1.0\columnwidth]{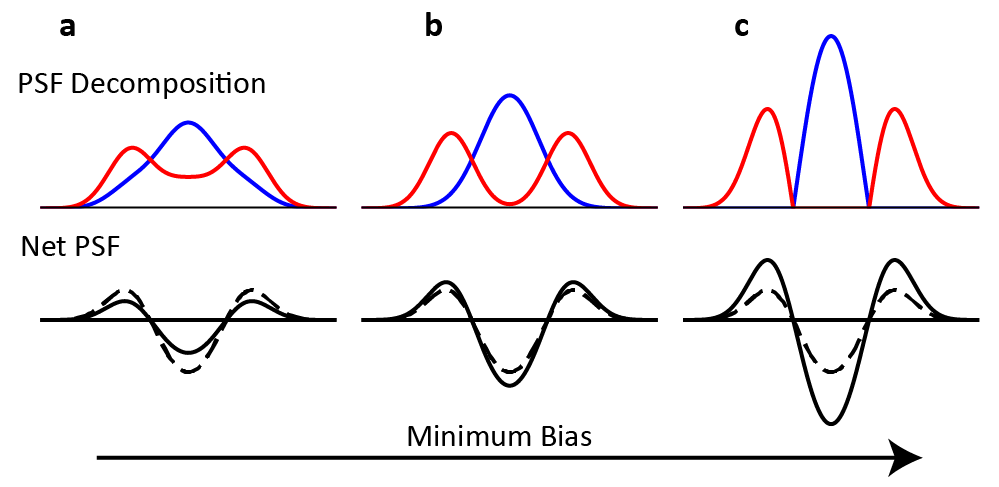}
    \caption{Depiction of the minimum-bias problem in multi-image-synthesis. On the first row, example decompositions with two component PSFs (red and blue) are shown for a Laplacian of Gaussian target. On the second row, the dashed black line corresponds to the target filter and the solid black corresponds to the synthesized net PSF. The computed mean signal-to-bias ratio (Equation \ref{eq:SBR}) from left to right are 0.38, 0.76, and 1.0. The decomposition in (c) corresponds to the minimum-bias solution.}
    \label{fig:minimum_bias_schematic}
\end{figure}

 While there are infinitely many solutions to this factorization problem, we highlight that not all will perform equally well in the presence of noise. Specifically, we consider a measurement model $\Gamma:\mathbb{R}\rightarrow\mathbb{R}$ mapping photons at the photosensor plane to detected electrical signal, where the noise scales with the signal intensity (see supplement S5 for details). The digitally-synthesized net PSF $\Gamma(H)\alpha$ may then be unusable if the net signal at a pixel is much weaker than the noisy component signals. This challenge of identifying optimal decompositions has historically been referred to as the minimum-bias problem \cite{Lohmann1978, Mait89}. We note that the consideration of measurement noise is also the reason that it is beneficial to optimize over all four polarization channels although one is not linearly independent. Specifically, it is the comparison between directly measuring $\Gamma(\psfD)$ as opposed to its digitally synthesized counterpart, $a_1\Gamma(\psfA) + a_2\Gamma(\psfB) + a_3\Gamma(\psfC)$ where $a_i$ are scalar constants (see Figure~\ref{fig:steerable_filtering} for a practical example). 

A qualitative example of different decompositions of varying quality are shown in Figure~\ref{fig:minimum_bias_schematic}. Optimal solutions to the \textit{unconstrained} problem may be characterized by orthogonality for the component signals that are to be digitally subtracted. To quantify the quality of a particular solution, the authors in \cite{Lohmann1978} propose as a metric the mean signal-to-bias ratio which may be given in a generalized form via, 
\begin{equation}
    \text{mSBR} = \left\| \lvert H\alpha\rvert \oslash H\lvert\alpha\rvert \right\| / N, 
    \label{eq:SBR}
\end{equation}
where $\lvert\cdot\rvert$ denotes an element-wise absolute value and $\oslash$ denotes Hadamard division. Throughout we apply vector-like norms for matrices $\left\|X\right\|_p = \left( \sum_{ij} \lvert X_{ij} \rvert^p \right)^{1/p}$, where $p=1$ if unstated. 

To identify a set of digital weights $\alpha$ and a metasurface $\Pi$ that together can realize target filtering operations, we then propose an optimization scheme utilizing gradient descent and a regularizer motivated by Equation (\ref{eq:SBR}). We formulate the objective as
\begin{equation}
    \operatorname*{argmin}_{\alpha,\Pi}  \mathlarger{\sum}_i \left[\left\| \frac{F^{(i)}}{\left\|F^{(i)}\right\|_2} - \frac{H\alpha^{(i)} }{\left\| H \alpha^{(i)} \right\|_2}  \right\|  + \mathcal{R} \right],
    \label{eq:objective}
\end{equation}
where the superscript $(i)$ enumerates over different sets of weights and targets. We use a two-term regularizer $\mathcal{R}$ given via,
\begin{equation}
    \mathcal{R} = -c_1\underbrace{\text{Tr}\left(R\right)}_\textsf{energy} + c_2 \underbrace{\left\| M\circ R \right\|}_\textsf{bias},
    \label{eq:regularizer}
\end{equation}
where $R = H^TH$, $M = \text{max}\left(-\alpha \alpha^T,0\right)$,
$c_1, c_2$ are hyper-parameters, and $\circ$ denotes the Hadamard product. 

Objective (\ref{eq:objective}) aims to synthesize net PSFs that approximate the set of target filters only up to scale. The (batched) matrix $R$ contains the terms $R_{i,j} = \langle h_i, h_j\rangle$ such that the elements on the diagonal are monotonically related to the energy in each polarization channel. The first regularizer term scaled by the coefficient $c_1$ then encourages the learned metasurface to have high focusing efficiency and the PSFs it induces to be spatially compact, i.e., contained within the finite simulation region. The second term controlled by the coefficient $c_2$ corresponds to a masked orthogonality constraint that minimizes the pair-wise overlap of PSFs with digital weights of opposite sign. In supplement S4, we show that this masked bias-regularizer emerges as a generalization of Equation (\ref{eq:SBR}) when considering distance metrics of the form $D\left( \lvert H\alpha\rvert, H\lvert\alpha\rvert \right)$. 

\begin{figure}[t!]
    \centering
    \includegraphics[width=1.0\columnwidth]{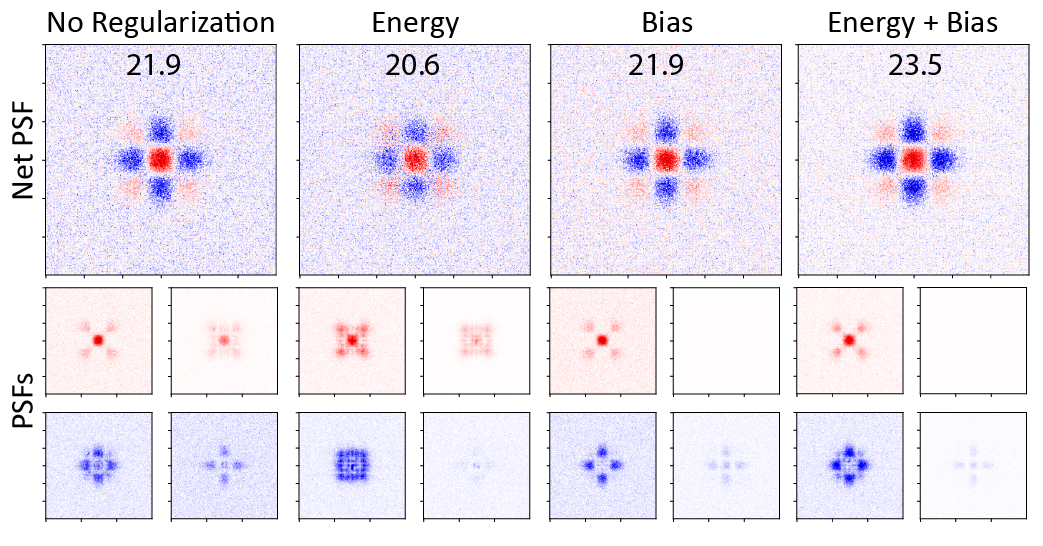}
    \caption{Visualization of the learned PSF decomposition for a metasurface optimized with and without regularization (monochromatic incident light and a single depth). The target filter is a second-derivative Gaussian kernel and a noisy measurement model $\Gamma(H)$ is applied to the PSFs. Overlaid text denotes the PSNR which here compares the similarity between the synthesized net PSF with noise and the target filter. Blue and red pixel colors correspond to negative and positive signal, respectively. The per-channel PSFs shown are displayed after the digital scaling, $\alpha_c\Gamma(h_c)$. When applying the bias regularization term, the gradient descent solution may learn to use fewer than all four images if beneficial via setting $\alpha_c$ terms close to zero.}
    \label{fig:objective_ablation}
\end{figure}
In Figure \ref{fig:objective_ablation}, we display an ablation study for the regularization terms (see supplemental Figure 8 for visualization with rendered images). Interestingly, for several target filtering kernels and initial conditions, we empirically observed that the unregularized gradient descent ($c_1=0, c_2=0$) naturally produced low-bias solutions but with a significant amount of energy deflected outside of the simulation region. Both the energy and bias regularization terms together were then required to achieve bias and energy efficient solutions. We note that while it is feasible to consider the application of noise via $\Gamma(\cdot)$ as a regularizer, we found that doing so produced unstable optimizations for noise levels that are large enough to have a substantial effect.

Lastly, we discuss an end-to-end variant of the objective in Equation (\ref{eq:objective}), used in this work to realize synthesized filters that operate under broadband illumination. We again define a target filtering kernel $F$ but we now compute the loss with respect to rendered images for planar scene radiances $\mathcal{I} \in \mathbb{R}_{\geq0}^{B\text{x}M\text{x}1}$ via,
\begin{align}
    \operatorname*{argmin}_{\alpha,\Pi}&  \mathlarger{\sum}_i \left[\left\| \frac{ F^{(i)} * \mathcal{I}}{\left\|F^{(i)}\right\|_2} - \frac{ (H*\mathcal{I})  \alpha^{(i)} }{\left\| H \alpha^{(i)} \right\|_2}  \right\|  + \mathcal{R} \right],
    \label{eq:broadband_objective}
\end{align}
where the spatial dimension of the tensors are unflattened prior to the 2D spatial convolution denoted by $*$. We note that it is not important here that the rendering treatment be accurate for complicated scenes. Rather, we leverage a loss based on convolved images in order to learn PSFs that yield an image transformation with similar statistics to the target operation. For example, while it is not possible to synthesize a compact net PSF that matches a fixed-width Laplacian of Gaussian kernel for all wavelength channels, we can instead discover a similar but physically realizable net PSF that approximates broadband edge-detection (see Section \ref{sec:broadband} for more discussion).

\begin{figure*}[ht!]
    \centering
    \includegraphics[width=1.0\linewidth]{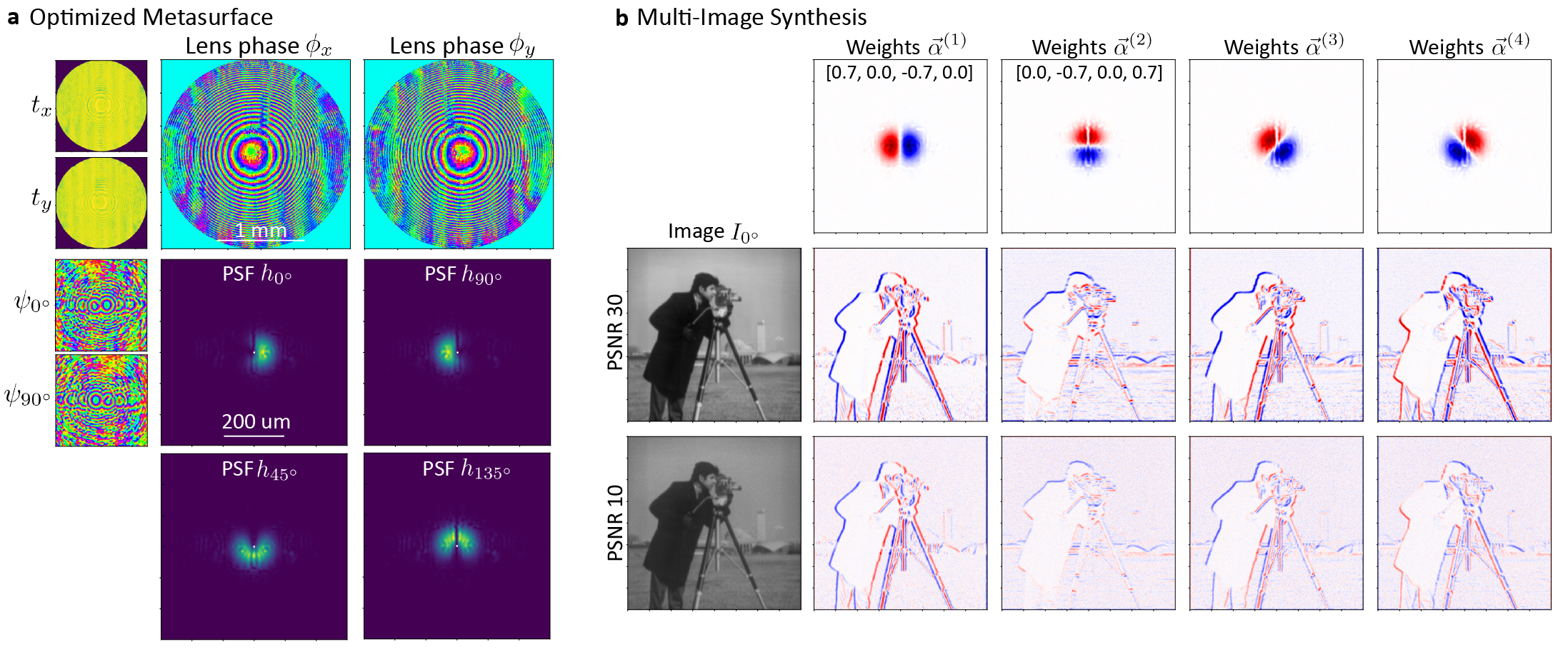}
    \caption{Multi-image synthesis for digitally steerable Gaussian first-derivatives. (a) The optimized metasurface phase and transmission imparted to incident light of two linear polarization states is shown (designed for infinity-focus). Below are the simulated PSFs for the four measurable polarization channels. Total focusing efficiency of the metasurface is approximately 71$\%$. (b) Top row displays the synthesized net PSFs where $\alpha^{(1),(2)}$ are learned and $\alpha^{(3),(4)}$ are defined by Equation (\ref{eq:steerable_weights}). Below are the synthesized, net images computed by first rendering the images for each polarization channel (where $I_{0^\circ}$ is displayed). The four images are then combined by per-pixel addition with weights $\alpha^{(i)}$.}
    \label{fig:steerable_filtering}
\end{figure*}

\section{Results}
In the following sections, we optimize $2$ mm diameter metasurfaces\footnote{While larger metasurfaces may be designed and fabricated, we choose $2$ mm optimizations as they can be done on a standard desktop GPU enabling easier accessibility.} according to objective Equations (\ref{eq:objective}) and (\ref{eq:broadband_objective}) and design multi-image synthesis systems for different tasks. Throughout we utilize an Adam optimizer with an exponentially decaying learning rate. All calculations are implemented in Tensorflow and we obtain gradients by automatic differentiation. When computing the PSFs, we evaluate the intensity and phase at the photosensor plane with sub-pixel resolution and use strided-convolutions to down-sample the field to match the simulated sensor's pixel pitch.   

In principle, the regularizer coefficients $c_1, c_2$ are hyper-parameters that should be chosen by a parameter sweep conducted for each task. In practice, however, we find that starting with reasonable initial conditions reduces the sensitivity to the exact values chosen. As an example, in Section \ref{sec:derivatives} where the target filter is a Gaussian derivative kernel, we initialize the metasurface to focus $\psfA$ and $\psfC$ to two off-axis points; in Section \ref{sec:broadband} where the target is edge-detection, we initialize to focus on-axis with different focal spot widths. In doing so, we find that we may set the bias regularizer coefficient $c_2$ to a single value that is fixed for all optimizations. We then conduct a coarse parameter sweep for the energy regularizer coefficient $c_1$ for each task, increasing the value and re-running the optimization until the total energy in the simulation region for optimized PSFs converged.  

When rendering images, the photosensor and its noise properties are modeled according to the EMVA standard \cite{emva1288} (see supplement S5 for details and the sensor parameters used). We specify the peak signal-to-noise ratio (PSNR) characterizing the simulated sensor noise, which is then used to scale the maximum brightness of the scene (number of photons) according to supplemental Eq. (3). While the PSFs are optimized over a smaller simulation region, the PSFs induced by the post-trained metasurfaces are computed across a larger area when used for rendering in order to capture the effects of scattered light. For demosaicing, we find it sufficient in most cases to use simple nearest-neighbor interpolation; we observe that the improvements from bi-linear interpolation or more advanced treatments are generally imperceptible since our target filters are relatively large and smoothly varying. 

We present experimental validation for the inverse-designed metasurfaces and the PSF decompositions in Section \ref{sec:experiment} (see also FDTD simulations in supplement S2). 

\subsection{Multi-filter Design: Steerable Derivatives}
\label{sec:derivatives}
We first demonstrate that it is possible to realize multiple opto-electronic filtering operations using a single fixed optic and the capture from a single exposure. Specifically, one may obtain different filtered images of a scene by changing only the digital synthesis weights $\alpha$. To do this, we exploit the class of steerable filters whose space of orientated kernels lie in the span of a small number of basis kernels, as discussed by the authors in \cite{Freeman1991}. In particular, we focus attention to the steerable Gaussian first derivative, parameterized by two basis kernels. Demonstrating examples that utilize a larger number of basis functions may be a topic of future investigation. The optical implementation is made possible by the fact that our architecture grants us access to four imaging channels, while we require at minimum only two channels per basis kernel in order to encode the positive and negative components of the signal.

In particular, for a co-designed set of PSFs $H$ induced by the metasurface, we desire a set of synthesis weights $\alpha^{(1)}$ that yields a net PSF corresponding to the Gaussian first derivative along the x-axis and another set of weights $\alpha^{(2)}$ corresponding to the first derivative along the y-axis. From this pair, the synthesis weights corresponding to the derivative along any other direction, specified by the rotation angle $\theta$, can be defined via,
\begin{equation}
    \alpha(\theta) = \alpha^{(1)}\cos(\theta) + \alpha^{(2)}\sin(\theta).
    \label{eq:steerable_weights}
\end{equation}
By optimizing for the two basis filters as targets using Equation (\ref{eq:objective}), we thus obtain an infinite (but compact) set of filters that can be digitally isolated. 

For simplicity, we design this metasurface to operate for monochromatic light of $\lambda=532$ nm and infinity-focus. The optical response of the optimized metasurface and its simulated performance in imaging a target scene is displayed in Figure~\ref{fig:steerable_filtering}. Notably, the optimization learns a minimum-bias decomposition to approximate the two target filters and the resulting metasurface can then produce differentiated images at any orientation angle. We also show that the synthesized filtered images are of suitable quality even when the component images are captured with low SNR.  

\subsection{Depth-dependent Differentiation}
We demonstrate that the filter-based optimization objective (Equation \ref{eq:objective}) can also be used to learn image transformations that are dependent on properties of the incident field. Specifically, we frame the target filter as a Gaussian first-derivative but with an orientation angle that varies with respect to the depth of an on-axis point-source. We then optimize for a metasurface $\Pi$ and a single set of digital weights $\alpha$. The synthesized image formed from this optical system would correspond to a differentiated image of the scene but with a spatially varying filter dependent on the depth of each object. 

\begin{figure}[t!]
    \centering
    \includegraphics[width=1.0\columnwidth]{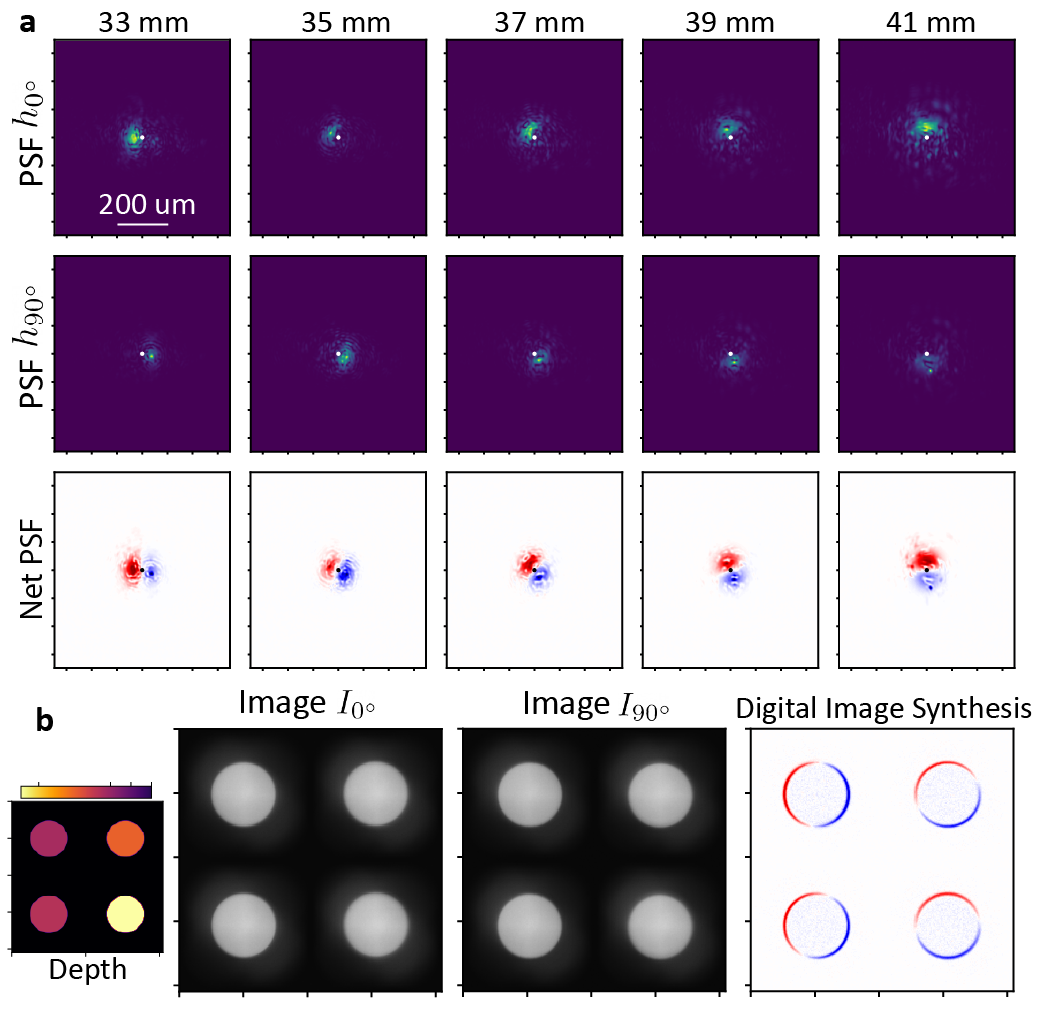}
    \caption{Multi-Image synthesis applied to the task of depth-dependent directional differentiation. The metasurface here is optimized for two-channel operation using $\psfA$ and $\psfC$ (a) The simulated PSFs and the synthesized net PSF are shown for point-sources at different depths. Total light efficiency is approximately 60$\%$. (b) The rendered component images and synthesized net image for a scene consisting of four uniform-intensity fronto-planar disks with relative depths indicated by the map (between 33 and 41 mm from the metasurface).}
    \label{fig:exp_depth_dependent_PSF}
\end{figure}

We consider monochromatic light of $\lambda=532$ nm and define the derivative orientation to vary linearly across a depth range of $1$ cm. Since a minimum of only two captured images are theoretically needed for this case, we conduct the optimization utilizing two polarization channels (trained by zero-masking the weight values $\alpha_{2,4}=0$). These results are shown in Figure~\ref{fig:exp_depth_dependent_PSF}a. We note that when the general four image case is considered with a non-zero bias regularizer, the optimized solution also converges to utilization of just two images. In either case, we find that the trained metasurface accurately learns to approximate the rotating kernel by encoding each lobe on orthogonal polarization channels. 

We also show in simulation the potential use of this optic for a simple test scene consisting of fronto-planar disks of uniform intensity at different depths, as displayed in Figure~\ref{fig:exp_depth_dependent_PSF}b. Inspired loosely by the principle of depth from differentiated images in an event-camera architecture \cite{Haessig2019}, we hypothesize that these depth-dependent derivatives may enable a unique approach to passive depth sensing by serving as a sparse depth cue that is coaligned with the undifferentiated, component images. Applying an equivalent spatially-varying kernel would be difficult to reproduce using a standard lens and digital post-processing. In the multi-image synthesis method, however, it emerges with a computational cost of as few as three floating point operations per pixel.

\subsection{Broadband Filter Design: Edge-detection}
\label{sec:broadband}
\begin{figure}[t!]
    \centering
    \includegraphics[width=1.0\columnwidth]{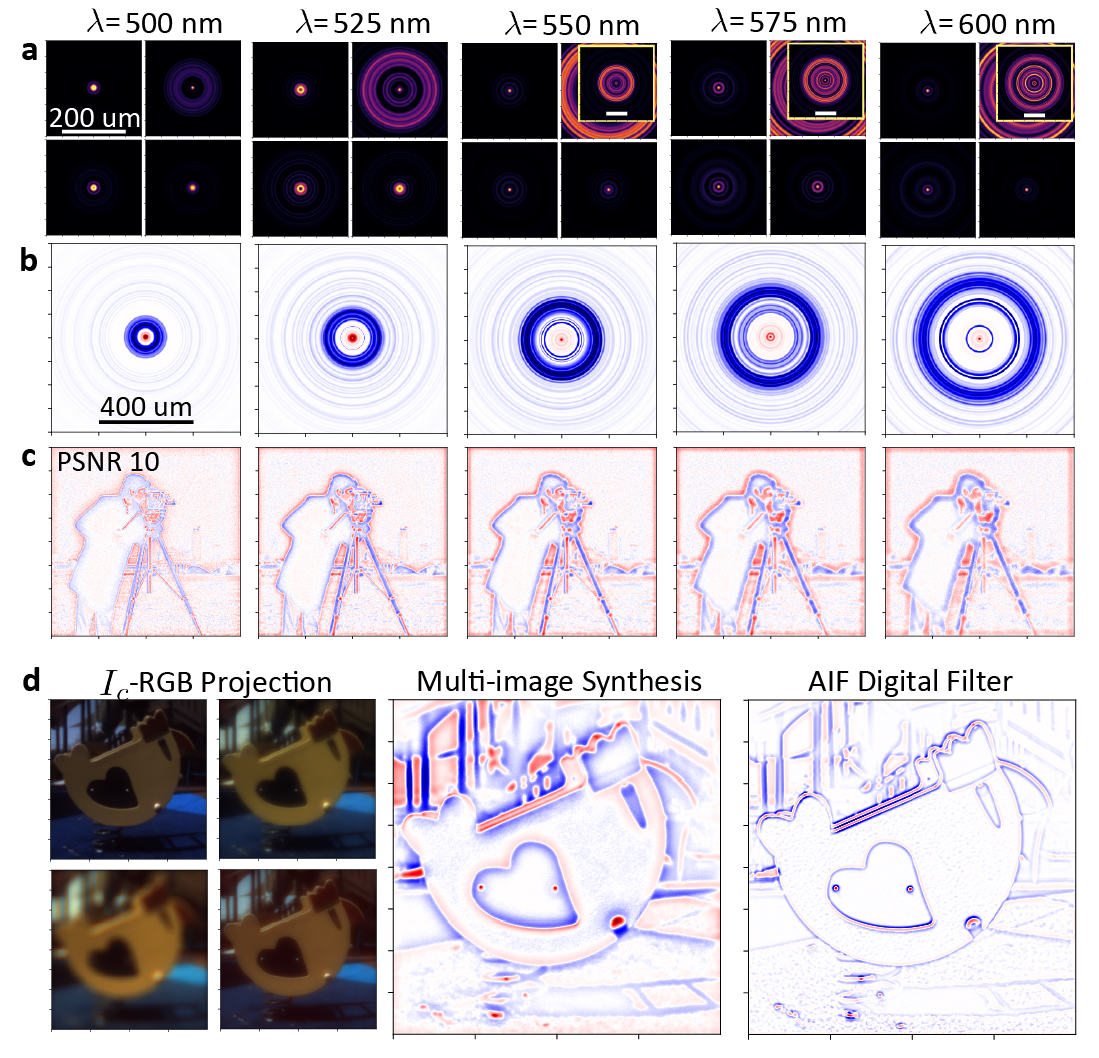}
    \caption{Broadband edge-detection (a) The PSFs for the four polarization channels at different wavelengths across the optimization range. The $\psfC$ intensities have a larger spatial extent and are shown on a larger grid in the overlaid insets. (b) The synthesized net PSFs for each wavelength band are displayed above (c) the corresponding per-wavelength band rendered image produced by convolution of the scene radiance and the net PSF (d) We utilize the post-trained metasurface and weights $\alpha$ to render images for different test scenes pulled from a hyperspectral dataset. While we consider a monochromatic photosensor for the synthesized images, on the left panel we project the broadband image at the photosensor for each polarization channel to RGB-color for visualization purposes.}
    \label{fig:broadband_MIS}
\end{figure}

Here we discuss the potential to leverage dispersion engineering in metasurfaces to the task of multi-image synthesis. We review that the PSFs produced by a metasurface vary with wavelength because both the optical response of cells (see Figure~\ref{fig:nanofin_MLP}b) and field propagation from the optic to the photosensor (see Equation \ref{eq:propagation}) are wavelength dependent. While we cannot control the latter, the freedom to select the cells that are placed at each location across the metasurface enables the ability to structure the PSFs with respect to wavelength. Importantly, the control and precision depends on the functional space of $t(\lambda)$ and $\phi(\lambda)$. Here, we continue utilizing nanofin cells; however, we highlight that a substantially larger design space can be realized by considering cells with more complicated nanostructures. An example is the three nanofin design introduced in \cite{Wchen2019}, which contains seven shape parameters per cell instead of two. Exploration of the filters that can be realized in such case is left to future investigations.

To demonstrate broadband capabilities, we utilize the image-based objective Equation (\ref{eq:broadband_objective}) and design an infinity-focused (radially symmetric) metasurface and a single set of synthesis weights. We optimize over a spectral range between $500$ and $600$ nm, with a $10$ nm step size. The target filter is defined to be a narrow Laplacian of Gaussian kernel and we utilize the ``camera man" image as the scene irradiance $\mathcal{I}$, both of which are kept the same for all wavelength channels. A key insight behind this approach is that we do not expect to discover a metasurface that realizes the user-defined filter \textit{exactly}; in fact, it is ensured that we cannot produce the wavelength invariant kernel specified here. By providing the filtered images as targets, however, we are able to find a decomposition that approximates the target statistics, which in this case are those characterizing edge-detection. We observe substantially better convergence by utilizing this approach rather than objective Equation (\ref{eq:objective}). 

\begin{figure}[t!]
    \centering
    \includegraphics[width=1.0\columnwidth]{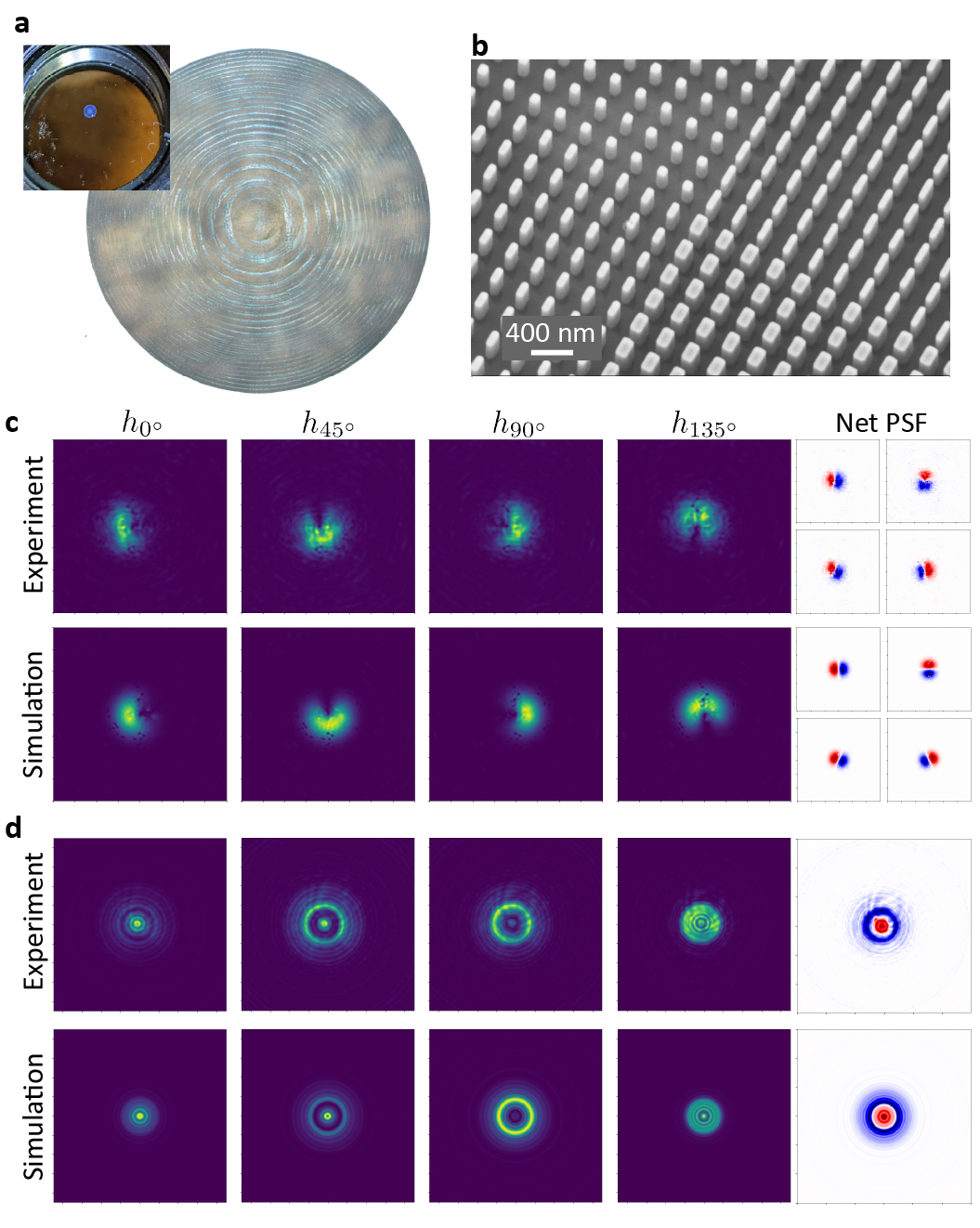}
    \caption{(a) Optical microscope image of the metasurface designed for steerable filtering. The inset shows a photograph of the mounted 2 mm device. (b) A scanning electron microscope image of a small region on the metasurface. The fabricated structures approximate the designed nanofins although machine limitations cause the rounding of edges. Experimental vs simulated PSFs and synthesized net PSFs are shown for (c) the steerable Gaussian first derivative kernel and (d) the Laplacian of Gaussian kernel, measured for incident light of $\lambda=532$ nm.}
    \label{fig:experimental_psfs}
\end{figure}

The results of this optimization are shown in Figure~\ref{fig:broadband_MIS}a-c. The four polarization channels are utilized and the learned synthesis produces a net PSF for each wavelength band with properties similar to the Laplacian of Gaussian kernel. The filtering operation in the post-trained system generalizes to other scenes, and in panel-\ref{fig:broadband_MIS}d (see also supplemental Figure 9), we display rendered synthesized images for test scenes utilizing the hyperspectral data released in \cite{NTIRE2022}. The spectral data is clipped to the optimization range, equivalent to assuming a wide pass-band spectral-filter at the entrance of the camera. The accuracy in which the synthesized images approximate the target spatial filtering operation may be improved by utilizing a collection of scenes during training rather than just one out-of-distribution scene.   

\subsection{Experimental Validation}
\label{sec:experiment}

\begin{figure}[t!]
    \centering
    \includegraphics[width=1.0\columnwidth]{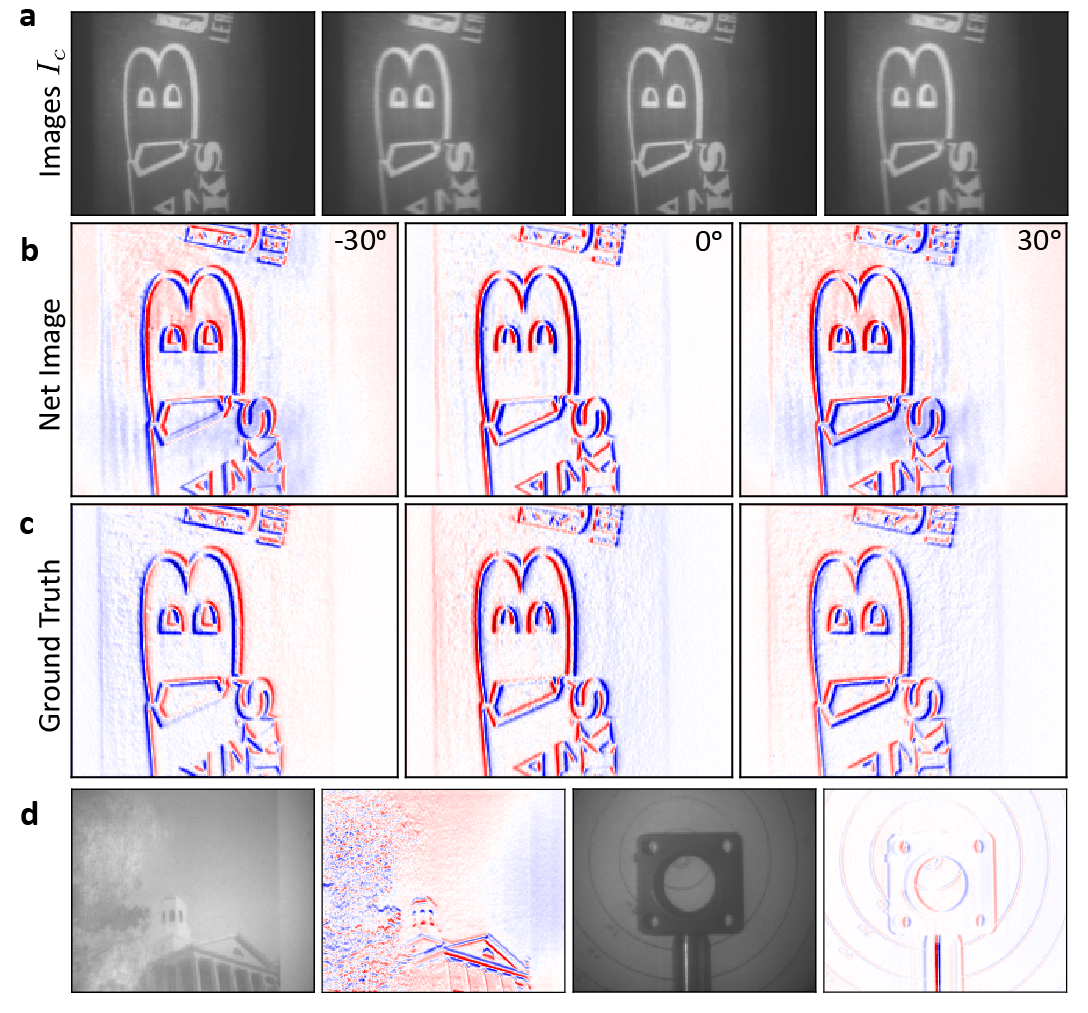}
    \caption{Images captured with the steerable-derivative metasurface camera. (a) Unprocessed measurements captured on the four polarization channels for a particular scene. (b) The net images formed from the pixel-wise linear combination of the four component images synthesizes differentiation that is digitally steered to three angles ($-30^\circ$, $0^\circ$, $30^\circ$) by only changing the summation weights $\alpha$. (c) Target filtering results at the same steering angles, computed by digital convolution of the target filters (Gaussian derivatives at orientation angles $0^\circ\pm 30^\circ$) and the in-focus raw image. (d) Measurements $I_{0^\circ}$ and the synthesized net image corresponding to differentiation along the x-axis for other scenes.}
    \label{fig:experimental_images}
\end{figure}
Lastly, we provide experimental validation for the design methods utilized in this work by fabricating and testing metasurfaces similar to those presented in Sections \ref{sec:derivatives} and \ref{sec:broadband}. We utilize electron-beam lithography and atomic layer deposition as discussed in \cite{devlin2016} to create the metasurface composed of 
 $600$ nm tall $\text{TiO}_2$ nanofins. Nanofabrication details are contained in supplement S6. Optical and scanning electron micrographs of one metasurface is displayed in Figure~\ref{fig:experimental_psfs}a-b, respectively. We then build an experimental camera utilizing an off-the-shelf polarizer-mosaicked photosensor (as shown in supplemental Figure 7) and simultaneously measure the four PSFs. The measurements are displayed in Figure~\ref{fig:experimental_psfs}c-d for monochromatic light of $\lambda=532$ nm, and we find good agreement between the experimental and simulated PSFs.

Utilizing the camera mounted with the steerable-derivative metasurface operating as a lens, we then capture images of various scenes, some of which are shown in Figure~\ref{fig:experimental_images}. Although the metasurface is designed to focus at infinity, we find good performance for objects placed as close as $1$~meter in front of the camera. By digitally computing only the weighted, pixel-wise summation of the four captured images, we confirm the ability to synthesize a collection of new, differentiated images of the scene with good agreement to the equivalent operations utilizing more expensive digital convolutions (shown in panels a-c).               

\section{Conclusion}
In this work, we have discussed the application of metasurfaces to the task of snapshot opto-electronic image processing. While the original theory introduced the principle of subtracting two normalized images, we present a generalization and a new design scheme for the learned linear synthesis of many images. Our experimental setup remains compact involving at minimum a single birefringent metasurface operating as a lens and a commercially available polarizer-mosaiced photosensor. By leveraging the unique properties of metasurfaces, we are able to demonstrate light-efficient polarization-encoded PSFs to realize multiple filters, along with depth-dependent and broadband operation. We also present a general discussion on the use of polarization for multi-coded imaging which may find use in other tasks beyond image processing. While orthogonal polarization states, e.g. $\psfA$ and $\psfC$, have been used before for other imaging tasks, we show that the intensity distributions formed from their interference may be engineered and utilized as additional imaging channels.    

Lastly, we highlight that metasurfaces have been used to produce multiple images by means other than polarization multiplexing. Guo et al. \cite{Guo2019} used a metasurface to produce two distinct images of a scene at spatially offset locations on the photosensor. By combining that method with the polarization technique discussed in this work, it is possible to capture \textit{eight} images of a scene in a single exposure. One may then optimize the image synthesis of all eight captures, producing a very large collection of different filters that can be isolated and applied with minimum computational cost. 



\ifpeerreview \else
\section*{Acknowledgments}
The authors thank Dr.~Zhujun Shi for many insightful discussions during the early stages of this work and the anonymous reviewers for their feedback and suggestions. This work was performed in part at the Harvard University Center for Nanoscale Systems (CNS). It was funded by NSF awards IIS-1900847 and IIS-1718012, and by NSF cooperative agreement PHY-2019786 (an NSF AI Institute, \url{http://iaifi.org}).\fi

\bibliographystyle{IEEEtran}
\bibliography{references}


\ifpeerreview \else

\begin{IEEEbiography}[{\includegraphics[width=1in,height=1.25in,clip,keepaspectratio]{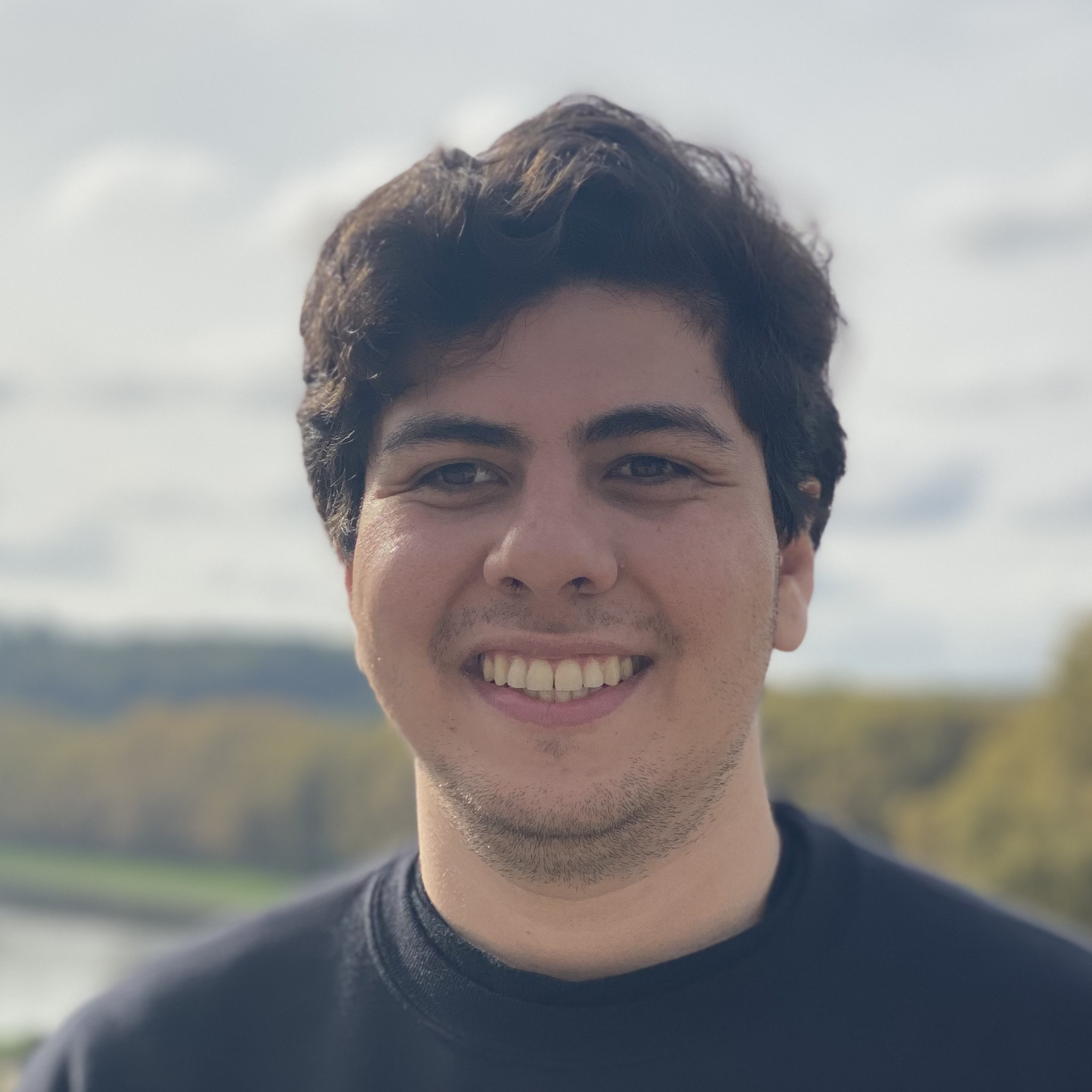}}]{Dean Hazineh} is a PhD student at Harvard University studying computational imaging and computer vision, advised by Todd Zickler and Federico Capasso. He is broadly interested in inverse problems and his research lies at the intersection of computer science, machine learning, and applied physics. 
\end{IEEEbiography}%

\begin{IEEEbiography}[{\includegraphics[width=1in,height=1.25in,clip,keepaspectratio]{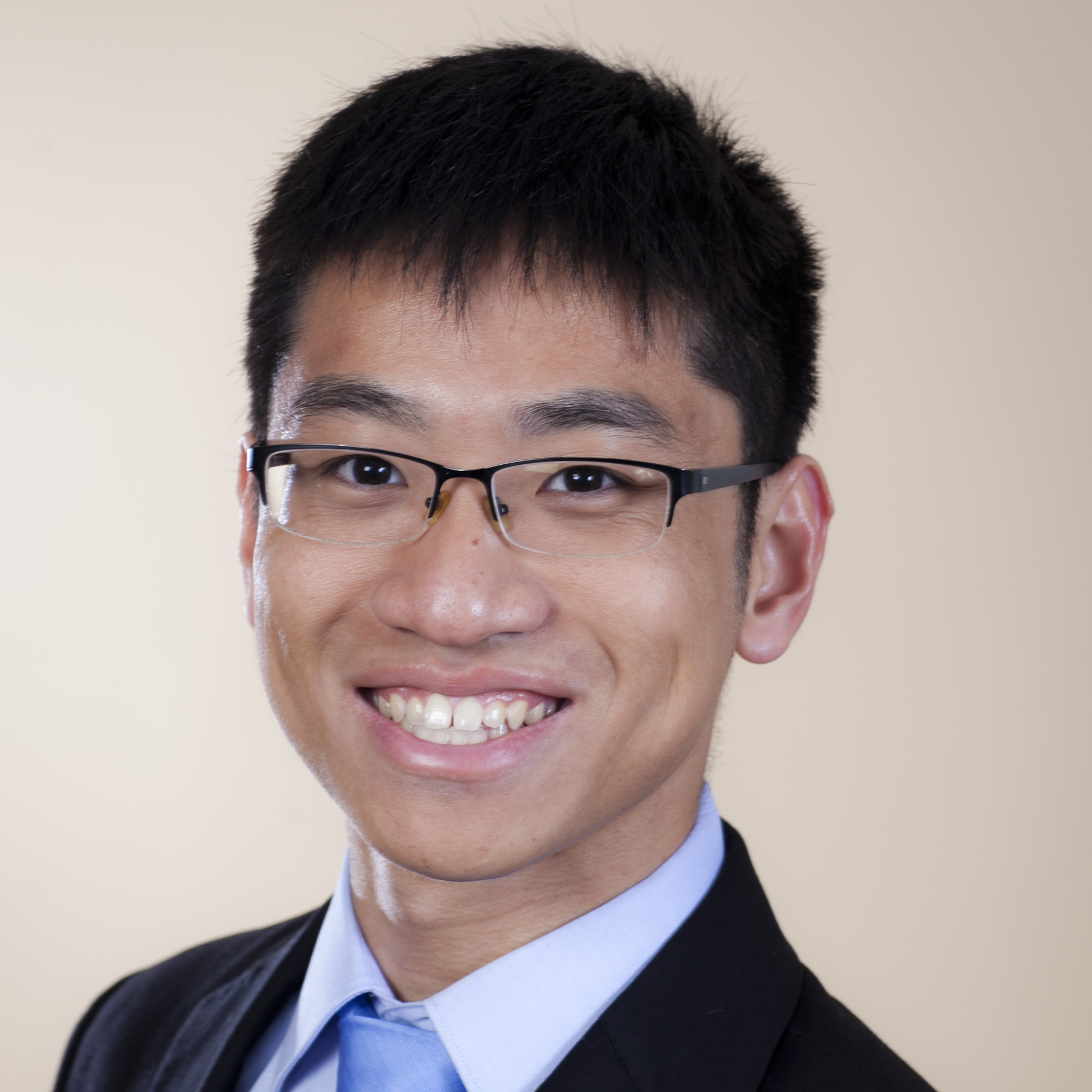}}]{Soon Wei Daniel Lim} is a PhD student at Harvard University and A*STAR National Science Scholarship fellow studying fundamental questions about structured light and darkness, topological inverse design, and nanostructured light-matter interactions. He is advised by Federico Capasso.
\end{IEEEbiography}%

\begin{IEEEbiography}[{\includegraphics[width=1in, clip, keepaspectratio]{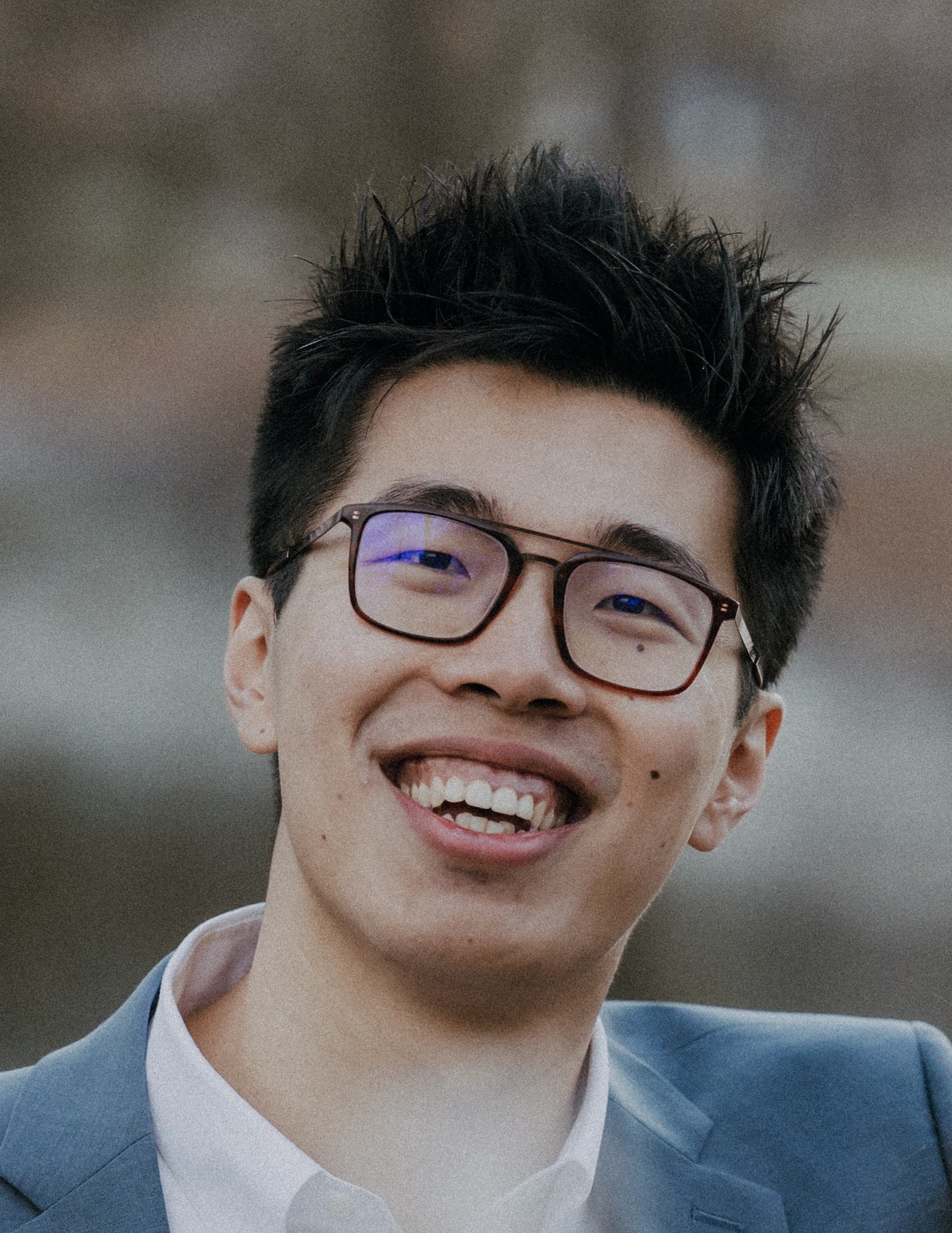}}]{Qi Guo} is an assistant professor at Purdue ECE. He innovates at the intersection of computer vision, machine learning, and optics to build next-generation low-power and compact visual sensors. Much of his and his collaborators’ research draws inspiration from biological eyes.  Dr. Guo received his Ph.D. degree from Harvard SEAS and B.E. in automation from Tsinghua. He and his coauthors received the Best Student Paper Award at ECCV 2016 and the Best Demo Award at ICCP 2018.
\end{IEEEbiography}%

\begin{IEEEbiography}[{\includegraphics[width=1in, clip, keepaspectratio]{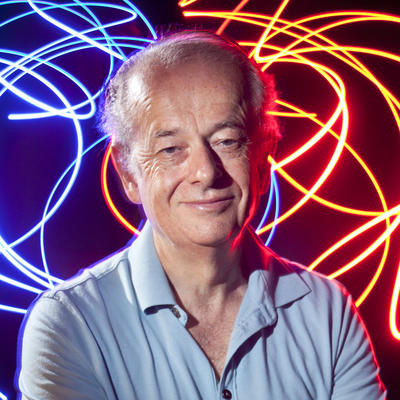}}]{Federico Capasso} is the Robert Wallace Professor of Applied Physics at Harvard University, which he joined in 2003 after 27 years at Bell Labs where his career advanced from postdoctoral fellow to Vice President for Physical Research. He has made contributions to optics and photonics, nanoscience, materials science, and QED, including the bandgap engineering technique leading to many new devices such as solid-state photomultipliers, resonant tunneling transistors and the invention of the quantum cascade laser. He and his group have carried out pioneering research on plasmonic and dielectric metasurfaces including the generalized laws of refraction and reflection, high performance metalenses and “flat optics”, and new methods to generate structured light. He carried out fundamental studies of the Casimir force, including new MEMS and the limits it places on this technology, and the first measurement of the repulsive Casimir force. He is a coauthor of over 500 publications and holds 70 US patents.
\end{IEEEbiography}%


\begin{IEEEbiography}[{\includegraphics[width=1in,height=1.25in,clip,keepaspectratio]{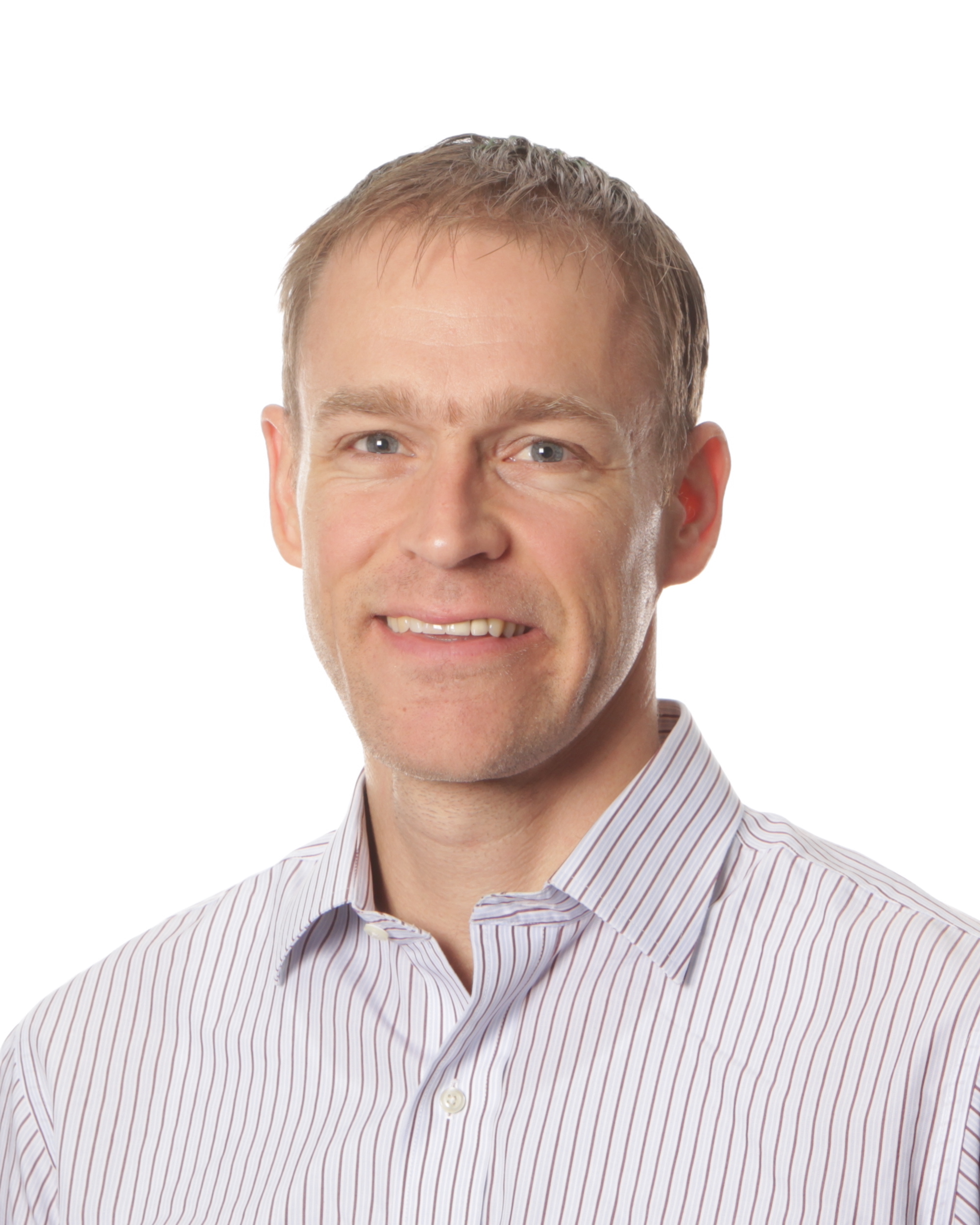}}]{Todd Zickler} received the B.Eng.~degree in honours electrical engineering from McGill University in 1996 and the Ph.D.~degree in electrical engineering from Yale University in 2004. He joined Harvard University in 2004, where he is now a professor of electrical engineering and computer science at the Harvard John A.~Paulson School Of Engineering And Applied Sciences. Zickler is grateful for having enjoyed sabbaticals as a visiting scientist at the Weizmann Institute of Science in Israel, a visiting scholar at Victoria University of Wellington in New Zealand, and a visiting professor at Kyoto University in Japan. His research models interactions between light, materials, optics and sensors, and it develops optical and computational systems to efficiently extract useful information from visual data. He is motivated by applications in robotics and augmented reality, and he is inspired by the intersections of computer vision, computer graphics, signal processing, applied optics, biological vision, and human perception. Zickler and his co-authors received the Best Student Paper Award at ECCV 2016, the Best Demo Award at ICCP 2018, and an honorable mention for the Best Student Paper Award at CVPR 2022. Zickler is a recipient of the National Science Foundation Career Award and a Research Fellowship from the Alfred P.~Sloan Foundation.
\end{IEEEbiography}%

\fi

\clearpage  

\ifarXiv
    

\section*{Supplementary material}

# Supplement: Polarization Multi-Image Synthesis with Birefringent Metasurfaces

Dean Hazineh, Soon Wei Daniel Lim, Qi Guo, Federico Capasso, Todd Zickler

◆

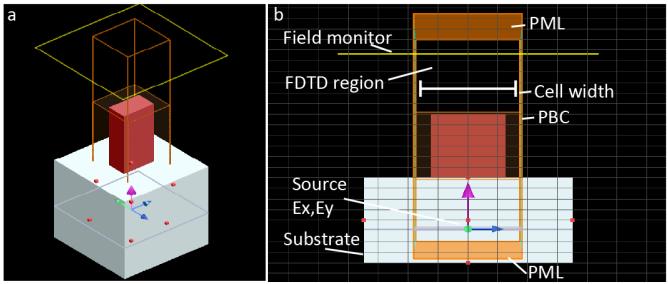

Fig. 1. Visualization of the cell FDTD calculation from a (a) 3D view and (b) 2D side-view. Field calculations are conducted on a grid of points within the FDTD region. PML refers to "perfectly matched layers" and PBC denotes periodic boundary conditions. We utilize the index of refraction for $SiO_2$ as the substrate and $TiO_2$ for the nanofin.

## S1. GENERATION OF THE CELL DATASET

The nanofin cell dataset utilized in this work was generated by finite-difference time-domain (FDTD) field calculations with the commercial software, Ansys Lumerical. We highlight for the interested reader that there are several other free code packages that could alternatively be used. Specifically, a different method to solve Maxwell's equations while assuming periodic boundary conditions that has gained significant attention recently in the metasurface community is the rigorous coupled wave analysis method (RCWA). Free and heavily validated RCWA code packages include [1], [2], [3].

A visualization of the FDTD calculations for a single cell instantiation is displayed in Fig. (1). The FDTD simulation region corresponds to a 3D (non-uniform) spatial grid of points, over which the electromagnetic fields are computed (displayed as the orange box in panels a-b). An ideal plane-wave source consisting of two linear polarization states, $E_x$ and $E_y$, is injected from within the substrate. Notably, periodic boundary conditions for the simulation region are used transverse to the incident light while perfectly matched layers are used at the top and bottom boundaries. The cell width, as noted in the main paper, corresponds here to the width of the FDTD simulation region.

While the fields are calculated throughout the simulation region, we collect the Fourier-transformed complex fields at the "monitor" (denoted by the yellow square in panels a-b). It is positioned a few hundred nanometers above the nanofin to avoid near-field effects. The 2D fields recorded

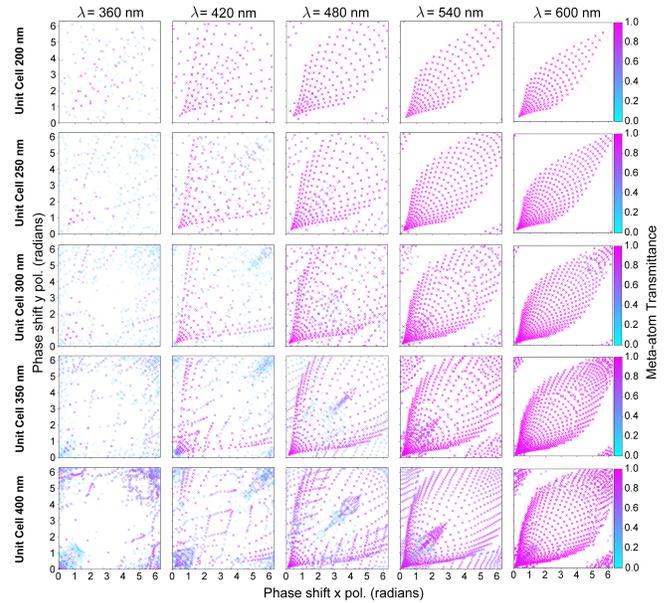

Fig. 2. Visualization of the nanofin optical response datasets for different cell widths (rows) and viewed for different wavelength slices (columns). The display is similar to Figure 1(b) of the main paper. Each point corresponds to a particular instantiation of the nanofin widths $w_x$ and $w_y$ but the point color (and transparency) map to the transmission percent. Complete coverage in the scatterplot would indicate that the library of cells can enable a decoupling of the $\phi_x$ and $\phi_y$ response.

at the monitor are then propagated to an axial point in the far field (a few micrometers above). We define the field at this centered, distant point to be the optical response of the cell. The amplitude transmittance and phase is normalized relative to the calculation with no nanostructure present.

In the main paper, we utilized a cell size of 350 nm; however, we also investigated the optical response dataset for other cell dimensions ranging from 200 nm to 400 nm. The results of this test when conducting a coarse sweep over nanofin widths are visualized in Figure 2. While the different datasets are qualitatively similar, we empirically observe that changing the cell size has an effect similar to scaling the amount of phase-delay imparted by a given nanofin. We chose the 350 nm cell width as it presented sufficient decoupling of the two phases $\phi_x$ and $\phi_y$ in the mid-visible near $\lambda = 530$ nm. A cell size of 400 nm could also be effective, however, we prefer selection of the smallest

usable cell dimension to avoid the potential for non-zero diffraction orders.

## S2. VALIDATION OF THE CELL DESIGN THEORY

In Section 3.1 of the main paper, we reviewed the cell design principle for metasurfaces. We consider the cells as independent building blocks and pre-compute their optical responses. By utilizing periodic boundary conditions in the calculations, we obtain an approximation to the true local optical response that is independent to the selection of nanostructures at other locations on the composite metasurface. Here, we demonstrate that this approximation is reasonable. We show that the PSFs computed for a metasurface using the cell model is almost equivalent to that obtained in the most general case where the field across the entire metasurface is solved for without partitioning.

We note, however, that it is generally computationally intractable to compute the fields across millimeter scale devices using a nanometer scale grid discretization. For this reason, we are only able to simulate the fields across 50 $\mu m$ diameter metasurfaces, using the same FDTD software as is discussed in supplement Section S1. Because we reduce the diameter of the metasurfaces relative to those considered in the main paper, we also reduce the lens-to-photosensor distances such that the designed f-number of each optic remains the same. This enables a better generalization of the findings.

For the FDTD calculations, we utilized 64 CPU cores and each full lens simulation took approximately 8 hours on a compute cluster. When calculating the fields across the full metasurface, we utilize perfectly matched layers for all the simulation boundaries (in contrast to the periodic boundary conditions that are used for the cell simulations). As a note, recent research on hardware and software acceleration has been leading to the development of specialized field solvers better suited to this task, a notable example is the recently released commercial software Tidy3D. In the future, simulations across millimeter or larger devices may become accessible.

We first review the analysis conducted for the task of steerable filters (similar to Figure 7 in the main paper). Utilizing the same optimization algorithm and target filter, we designed a 50 $\mu m$ metasurface for infinity focus and monochromatic incident light of $\lambda = 532$ nm. The optimized arrangement of nanofins can be seen in supplemental Figure 3a. In panel (b), we show the computed, modulated fields that are transmitted through the full metasurface in response to a normally-incident, linear polarized plane-wave, whose polarization angle is orientated at $45°$ with respect to the x-axis. The FDTD calculations are computed for the full metasurface without partitioning while the cell-based treatment stitches together the predicted, spatial modulation pattern based on the pre-computed dataset of cell optical responses. Notably, we observe excellent agreement in the phase predictions for the metasurface when assuming the cell-based treatment vs the more general but expensive full model. Qualitatively, the transmittance also has strong agreement although we observe more variations.

We now compare the predicted PSFs for both cases. The PSF calculations for the cell-based approach is done in the

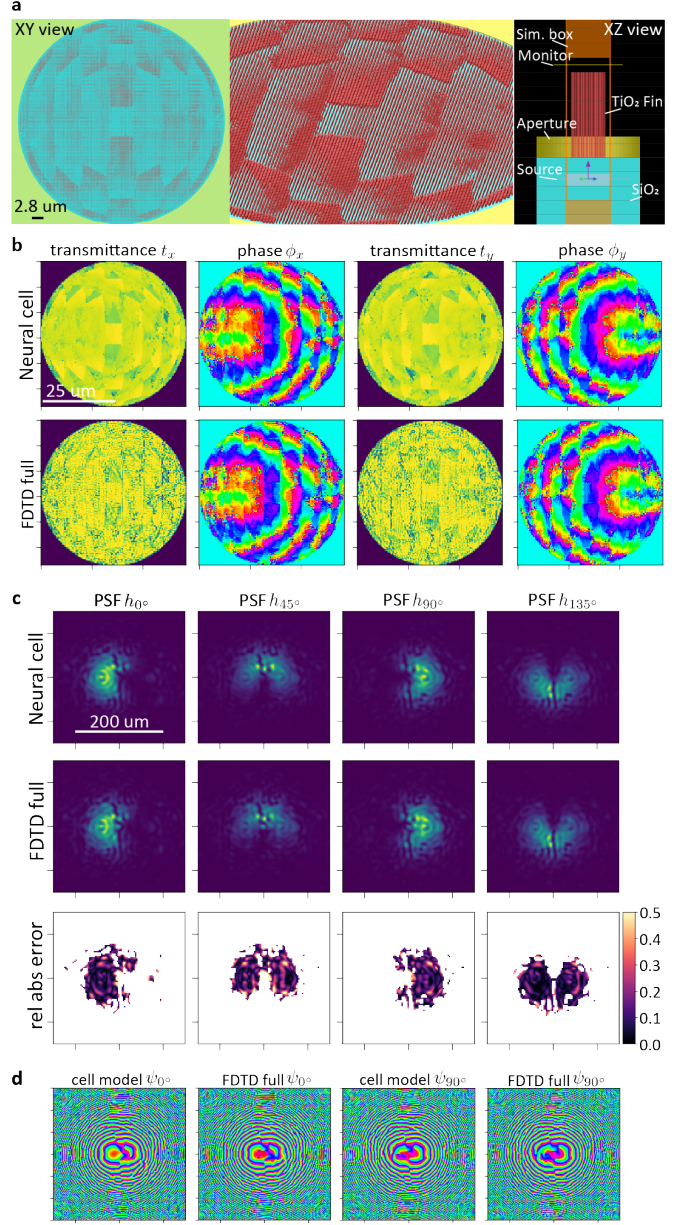

Fig. 3. A comparison of full-field FDTD calculations to the cell-based calculations presented in the main paper for a reduced size metasurface that implements the steerable Gaussian derivatives. (a) The optimized 50 $\mu m$ diameter metasurface loaded into the FDTD software. Details of the calculation are shown in the right-most panel. (b) The phase and transmittance just after the metasurface, in response to a normally incident plane wave of wavelength $\lambda = 532$ nm. We refer to the mapping from the nanofin cell to the predicted optical response when utilizing the pre-trained MLP as the "neural cell" prediction. FDTD full refers to the direct calculation of the field when simulating the entire device. (c) Predicted intensity PSFs from the cell-based treatment and Fresnel propagation vs directly from FDTD. The relative absolute error is computed and shown only for pixels with an intensity of at least 5% the peak intensity. (d) The phase distribution at the output plane computed by both methods.

same manner as in the main paper. We first utilize the pre-trained MLP to map the metasurface cells to their local optical response. We then propagate the field defined by the collection of per-cell responses to the photosensor plane using the Fresnel integral. We do this calculation directly for

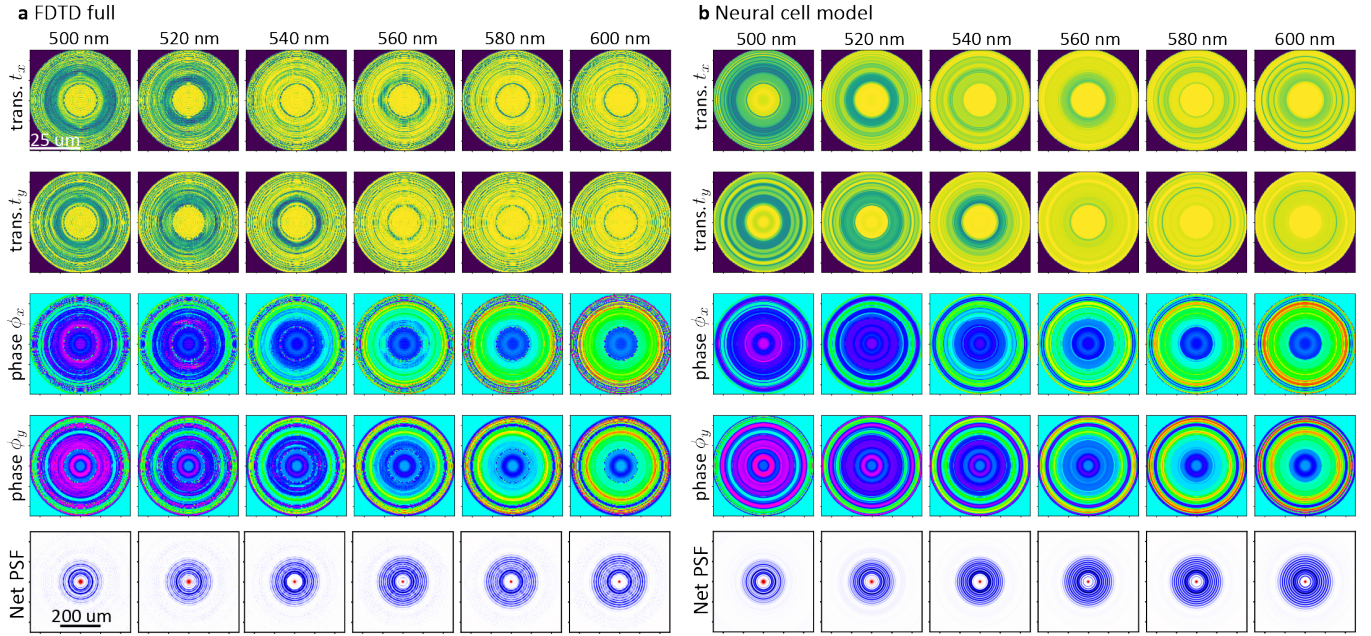

Fig. 4. Similar to supplemental Figure 3, we compare the cell-based approach (and the Fresnel propagated PSFs) to the equivalent calculation utilizing FDTD for the $50~\mu m$ metasurface shown in supplemental Figure 5.

$h_{0°}$ and $h_{90°}$ and then obtain $h_{45°}$ and $h_{135°}$ by computing the interference. This set of PSFs are shown in supplemental Figure 3c-d.

For the full FDTD case, we first directly solve for the field after the metasurface. We then use the FDTD software itself to propagate this field to the output plane; in doing so, a rigorous treatment of propagation is used which differs from the Fresnel integral and does not assume the paraxial approximation. This comparison thus also provides validation for our differential propagator and treatment of interference. The FDTD predicted PSFs are also displayed in panel c-d, and we find excellent agreement between the cell based calculations and the more rigorous FDTD treatment.

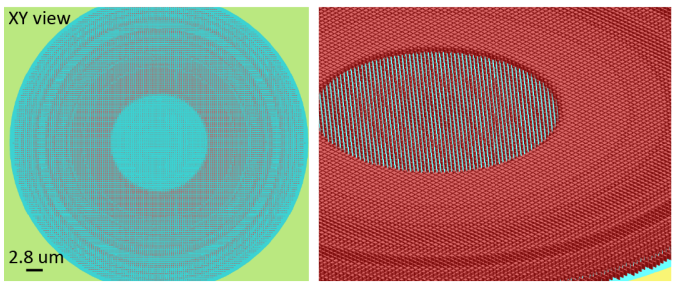

Fig. 5. Visualization of the optimized metasurface to be used in a broadband setting. This metasurface corresponds to the analyzed profiles in supplemental Figure 4.

We now explore the full-field analysis applied to a broadband case, shown in supplemental Figure 4. The technical details of the cell model and FDTD calculations mirror the above discussion, so we focus here on the results. Similar to Section 4.3 of the main paper, we consider the design of a metasurface that is used to approximate the Laplacian of Gaussian kernel for a wide range of incident wavelengths. For this design task, however, we find that it is suitable to apply the filter-based objective given by Equation (10) in the main paper. We define the target filter kernel to have a width that increases with wavelength which matches the natural broadening of the PSF. The optimized nanofin metasurface is shown in Figure 5. Again, we compare the FDTD calculation (left panel of supplemental Figure 4) to the cell-based approach (right panel) and find strong agreement for the predicted transmittance, phase, and PSFs across the wavelength range.

## S3. NEURAL OPTICAL MODEL EVALUATION

For gradient-based optimization of the metasurface, we require an efficient and differentiable approximation for the mapping between the nanostructure shape parameters and the optical response of the cell. As an alternative to the MLP, we also considered for this work elliptic radial basis function networks (ERBFs) and simple multivariate polynomial functions as applied in [4]. For all cases, we consider the input-output mapping depicted in Figure 2a of the main paper. To the best of our knowledge, ERBFs have not previously been explored in the context of this problem. First we review the performance of these alternative representations and after, we discuss auto-differentiable field solvers as a benchmark.

ERBF networks are reviewed in [5], [6] and may be considered as a particular class of neural networks consisting of a single hidden layer. Each neuron in the hidden layer parameterizes a radial basis function, which in this case corresponds to a three-dimensional elliptic Gaussian potential (assuming the set of three inputs $w_x$, $w_y$, and $\lambda$). The neurons then compute the Euclidean distance of the inputs and its *weights* followed by the activation, in contrast to neurons in a typical MLP which utilize the dot-product between inputs and weights. The number of neurons in the

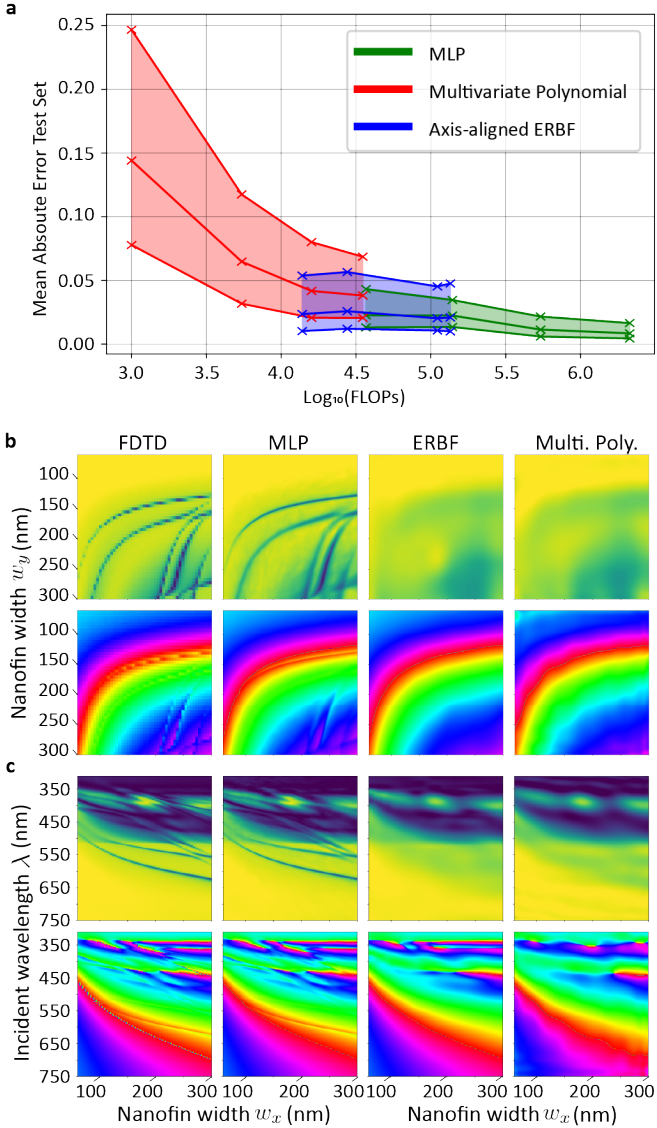

Fig. 6. A comparison of the MLP to other implicit representations. (a) We train several models of different sizes and evaluate the performance of each on a test-set of nanofin cells. In each case, we compute the predicted complex optical response for two basis polarization states and evaluate the mean absolute error relative to the true optical response. FLOPs refers to the number of floating point operations required per model evaluation. (b-c) This figure is analogous to Figure 2b-c of the main paper. We display the model predictions for slices through the dataset at (b) a fixed wavelength of $\lambda = 532$ nm and (c) at a fixed nanofin width $w_y = 180$ nm. The points are queried at 5x the resolution of the training data.

networks required a similar number of floating point operations (FLOPs) as the MLP but took orders of magnitude longer to train.

Alternatively, the multivariate polynomial formulation is relatively standard and the mapping is cast into a linear, matrix form. The coefficient matrix is updated by the method of least squares and we consider a separate matrix for each of the six output parameters. We tested polynomial orders up to 24, above which became impractical due to memory limitations. We also chose not to explore higher polynomial orders utilizing a stochastic gradient descent training scheme.

The test performance and computational complexity defined by the number of floating point operations is displayed in Figure 6 for the different models. For the MLP and the ERBF networks, the same set of test cells in the dataset were withheld during training. We find that the MLP substantially outperforms the other two representations in terms of achievable accuracy. While the largest MLP (1024 neurons in each hidden layer) required an order of magnitude more FLOPs per inference than the other largest models tested, this difference proved unimportant as the MLP is still efficient enough to be used in the optimization of a 2 mm metasurfaces with a single desktop GPU (RTX 3090). Moreover, the smallest MLP (e.g. 256 neurons) also outperforms the other representations. The model predictions when reproducing slices of the dataset are visualized in panels b-c.

Lastly, we compare the computational complexity of these approximate models to the direct field calculations evaluated utilizing an auto-differentiable field solver. Here, we consider the Tensorflow implementation of rigorous coupled wave analysis (RCWA) released in [2]. RCWA solves Maxwell's equations in the Fourier domain and formulates the problem as an eigenequation. For small cell sizes and periodic boundary conditions, RCWA is generally more computationally efficient than FDTD. The accuracy of the calculations depends on the grid discretization used when defining the simulation and on the number of Fourier modes applied to the solution; typically 81 or more Fourier modes are required to obtain converged results.

For this study, we consider a discretization of the nanofin cell into a 512x512 grid. We then consider 49, 81, and 121 Fourier modes. Using the Tensorflow profiler, we then estimate the number of floating point operations required per RCWA cell evaluation (and per wavelength) to be approximately, 363M, 1.62B, and 5.38B. Although memory bottlenecks are generally the limiting factor for applying auto-differentiable field solvers to inverse design problems, we summarize that RCWA would be several orders of magnitude more expensive as compared to the MLP from the perspective of number of FLOPs.

## S4. MINIMUM-BIAS REGULARIZATION

We now discuss the motivation for the particular form of the bias regularization term introduced in Equation (11) of the main paper. The inspiration comes from analysis of the per-pixel "signal-to-bias" ratio metric introduced by the authors in [7], which we have rewritten in a generalized form in Equation (9) of the main paper. Since computing

hidden layer and thus the number of basis functions used to represent the data is a degree of freedom. The neuron's standard deviations and center coordinates, in addition to the weights and bias of a linear output projection layer, are all trainable parameters that may be updated by stochastic gradient descent.

We tested axis-aligned ERBFs with a number of neurons between 512 and 5000 to represent the nanofin cell dataset. Increasing the number of neurons beyond this point became impractical from a training time perspective. While we also explored the more general case of learnable covariance matrices rather than axis-aligned standard deviations, these

the per-pixel ratio may be numerically unstable, we instead consider differential distances $D(\cdot)$ between the two vectors/matrices:

$$D(|H\alpha|, H|\alpha|) \equiv D\left(\left|\sum_c \alpha_c h_c\right|, \sum_c |\alpha_c| h_c\right), \quad (1)$$

where we are utilizing the notation $|X| = \sqrt{X \circ X}$ to denote a per-element absolute value utilizing the Hadamard product $\circ$, $c$ enumerates the four polarization channels $c \in \{0°, 45°, 90°, 135°\}$, $\alpha_c$ denote a scalar constant used in the digital synthesis, and $h_c$ is a PSF from the set.

We find that it is insightful to first consider the difference $d$ of the *squared* vectors[1] which may be expanded via:

$$\begin{aligned}
d &= \left(\sum_c \alpha_c h_c\right)^2 - \left(\sum_c |\alpha_c| h_c\right)^2 \\
&= \sum_i \sum_j \left(\alpha_i \alpha_j - |\alpha_i||\alpha_j|\right) h_i \circ h_j \\
&= 0 + \sum \sum_{i \neq j} \left(\alpha_i \alpha_j - |\alpha_i||\alpha_j|\right) h_i \circ h_j \\
&= \sum \sum_{\text{sign}(i) \neq \text{sign}(j)} -2|\alpha_i||\alpha_j| h_i \circ h_j.
\end{aligned}$$

From the last line, we then observe that the L1-norm of this difference vector $d$, $\|d\|_1$, corresponds to the regularization that is used in the main paper. Specifically, summing over elements of $d$ yields the overlap integral for PSFs that are combined with digital weights of opposite sign.

We now show that this masked regularization occurs up to a normalization term when considering the non-squared vector difference for $d$. In other words, variations of the masked orthogonality emerge when $D$ in Equation (1) corresponds to standard vector norms:

$$\begin{aligned}
d &= \left|\sum_c \alpha_c h_c\right| - \sum_c |\alpha_c| h_c \\
&= \frac{(\sum_c \alpha_c h_c)^2 - (\sum_c |\alpha_c| h_c)^2}{|\sum_c \alpha_c h_c| + \sum_c |\alpha_c| h_c} \\
&= \frac{\sum \sum_{\text{sign}(i) \neq \text{sign}(j)} -2|\alpha_i||\alpha_j| h_i \circ h_j}{|\sum_c \alpha_c h_c| + \sum_c |\alpha_c| h_c}
\end{aligned}$$

In the last line, the denominator is a purely positive vector. If we compute $\|d\|_1$ in this case, we again obtain the overlap integral for pairs of PSFs but with a per-pixel weighting dependent on the intensity distributions.

## S5. THE MEASUREMENT OPERATOR AND SIMULATED NOISE

As introduced in Section 3.4 of the main text, we define a noisy measurement operator $\Gamma(\cdot)$ which maps the photons at the photosensor plane to detected electrical signal. This measurement model is defined according to the EMVA standard [8] via,

$$\Gamma(X) = \text{Round}\left[(\mathcal{P}(qX) + \mathcal{N}(\mu_d, \sigma_d)) k\right], \quad (2)$$

1. While this type of vector difference is not used in standard distance metrics, it is helpful to consider as a start given the abnormal usage of the absolute value

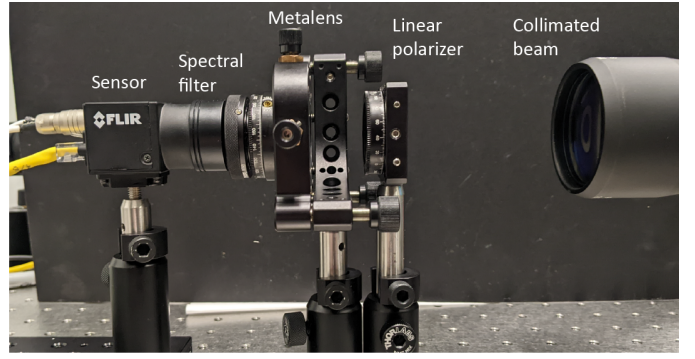

Fig. 7. Experimental setup used to measure the polarization encoded PSFs, discussed in Section 4.4 of the main paper. The polarization camera used is the FLIR Blackfly BFS-PGE-51S5; the global coordinate frame is defined relative to the photosensor, such that the x- and y-axis directions aligns with the $0°$ and $90°$ nanowire polarizers on the detection pixels. The linear polarizer at the entrance of the optical system is orientated at $45°$.

where $\mathcal{P}$ denotes the Poisson and $\mathcal{N}$ the normal distribution, $q$ and $k$ are the detector quantum efficiency and gain, and $\mu_d$ and $\sigma_d$ parameterize the dark noise. For the simulations in this work, we model the noise properties of the BFS-PGE-51S5 photosensor, which is the polarizer-mosaicked photosensor used in the experiment (and shown in supplemental Figure 7). EMVA technical specifications for this photosensor are available from the manufacturer (although reported only for $525$ nm incident light); we utilize the values $q = 0.24$, $\mu_d = 2.45 \ e^-$ and inverse gain $k^{-1} = 0.18$ for all wavelengths and polarizations.

When rendering scenes via Equation (2) of the main paper, we desire a rescaling of the scene irradiance $\mathcal{I}_c(u, v)$ such that the per-channel intensity produced at the photosensor by an ideal lens corresponds to a noisy measurement with a specified peak signal-to-noise ratio (PSNR). We identify this scaling factor using the following relation [8]:

$$N_{\text{photons}}(\text{PSNR}) = \frac{\text{PSNR}^2}{2q}\left(1 + \sqrt{1 + \frac{4(\sigma_d^2 + \sigma_q/k^2)}{\text{PSNR}^2}}\right). \quad (3)$$

Here, $N_{\text{photons}}$ refers to the peak number of photons which sets the maximum value of $\mathcal{I}_c$. The variable $\sigma_q$ corresponds to quantization noise which in this case is defined by a 12 bit ADC conversion. By utilizing supplemental equations (2)-(3), we are able to predict the effectiveness of designed multi-image synthesis systems which are generally well-known to be sensitive to measurement noise.

## S6: METASURFACE NANOFABRICATION

The metasurface design is written into 600 nm thick ZEP520A positive electron beam resist (Zeon Specialty Materials Inc.) using electron beam lithography (Elionix HS-50 50 kV). The resist voids are back-filled with amorphous titanium dioxide using atomic layer deposition (Cambridge NanoTech Savannah) and the excess titanium dioxide is etched back using inductively-coupled plasma reactive ion etching (Oxford PlasmaPro 100 Cobra). The electron beam resist is removed by overnight immersion in Remover PG (Kayaku Advanced Materials). An opaque gold mask (2

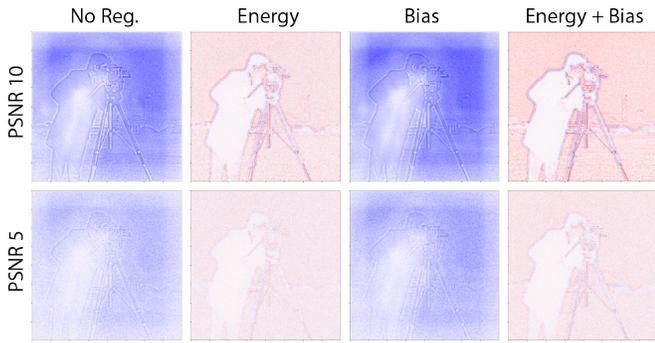

Fig. 8. Similar to the ablation study presented in Figure 6 of the main text. Here, we compare the image synthesis performance of optimized metasurfaces when designed with and without the regularization terms. For visual simplicity, we target a symmetric Laplacian of Gaussian kernel. Images are rendered using PSFs computed over a larger simulation area than that used during optimization (see Section 4 introduction for more details). Regularization has a substantial effect and the energy term enforces a spatially compact PSF. When paired with the bias term, the quality of synthesized images are improved. The importance of minimum-bias solutions can diminish with increasing SNR.

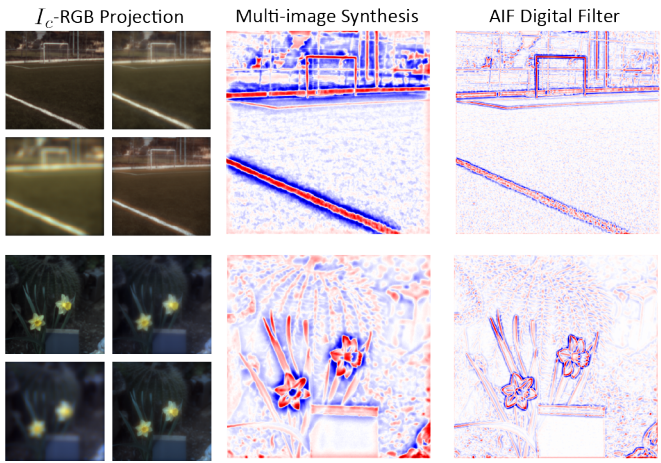

Fig. 9. Additional examples of the broadband, rendered and synthesized images produced by the optimized nanofin metasurface, equivalent to Figure 9d in the main paper but for different test scenes. See the caption and text for more details.

mm diameter) is deposited around the metasurface by positive tone photolithography in S1813 photoresist (Kayaku Advanced Material) with direct laser writing (Heidelberg MLA150), followed by electron beam evaporation of 5 nm of chromium and 200 nm of gold (Denton E-beam Evaporator). Residual photoresist is removed by overnight immersion in Remover PG (Kayaku Advanced Materials).


\fi

\end{document}